\documentclass[eng,article,submit,moreauthors]{Definitions/mdpi}
\usepackage{mathtools,longtable,algorithm,algpseudocode,placeins,listings}
\usepackage{expl3}
\ExplSyntaxOn\pdf_version_gset:n {1.7}\ExplSyntaxOff
\let\originalincludegraphics\includegraphics
\renewcommand{\includegraphics}[2][]{%
  \IfEndWith{#2}{.eps}{%
    \StrSubstitute{#2}{.eps}{.pdf}[\convertedgraphic]%
    \originalincludegraphics[#1]{\convertedgraphic}%
  }{\originalincludegraphics[#1]{#2}}}
\lstdefinestyle{referenceprompt}{basicstyle=\ttfamily\footnotesize,breaklines=true,columns=fullflexible,keepspaces=true,showstringspaces=false,frame=single,aboveskip=8pt,belowskip=8pt}
\AtBeginEnvironment{equation}{\linenomath}
\AtEndEnvironment{equation}{\endlinenomath}
\AtBeginEnvironment{align}{\linenomath}
\AtEndEnvironment{align}{\endlinenomath}
\AtBeginEnvironment{align*}{\linenomath}
\AtEndEnvironment{align*}{\endlinenomath}

\newcommand{\artifacturl}{\url{https://github.com/Tt200411/From-Fixed-Defaults-to-Calibrated-Activation-Policies-for-Long-Term-Time-Series-Forecasting}}
\newcommand{\artifactlink}{\href{https://github.com/Tt200411/From-Fixed-Defaults-to-Calibrated-Activation-Policies-for-Long-Term-Time-Series-Forecasting}{GitHub repository}}
\firstpage{1}\pubvolume{1}\issuenum{1}\articlenumber{0}
\datereceived{ }\daterevised{ }\dateaccepted{ }\datepublished{ }\hreflink{https://doi.org/}
\pubyear{2026}\copyrightyear{2026}
\Title{From Fixed Defaults to Calibrated Activation Policies for Long-Term Time-Series Forecasting}
\TitleCitation{Calibrated Activation Policies for Long-Term Forecasting}
\Author{Xu Shijie $^{1,3,\dagger}$, Boyan Tang $^{2,3,*,\dagger}$, Kai He $^{5}$ and Wu Jianghua $^{3,4}$}
\AuthorNames{Xu Shijie, Boyan Tang, Kai He and Wu Jianghua}
\AuthorCitation{Xu, S.; Tang, B.; He, K.; Wu, J.}
\address{%
$^{1}$ \quad School of Automation, Nanjing University of Information Science and Technology; 202412232330@nuist.edu.cn\\
$^{2}$ \quad School of Engineering, Hong Kong University of Science and Technology;\newline btangan@connect.ust.hk\\
$^{3}$ \quad Shenzhen Research Institute of Big Data, The Chinese University of Hong Kong,\newline Shenzhen, Guangdong, China;\newline wujianghua@sribd.cn\\
$^{4}$ \quad Shenzhen Loop Area Institute, Shenzhen, Guangdong, China\\
$^{5}$ \quad School of Aerospace, Mechanical and Mechatronic Engineering, The University of Sydney;\newline kahe0116@uni.sydney.edu.au}
\corres{Correspondence: btangan@connect.ust.hk (B.T.)}
\firstnote{These authors contributed equally to this work and share first authorship.}
\abstract{Long-term time-series forecasting supports decisions in energy, transportation, and other applications where reliable predictions are needed beyond the observed data. Although forecasting research has advanced model architectures, fixed activation configurations provide limited guidance on how activation choices should vary across datasets and prediction horizons. This study aims to establish when task-specific activation choices improve forecasting accuracy and how previous experiments can guide the selection of suitable configurations for new tasks. We develop a calibrated selection policy that learns a reliable default from prior experiments, adapts the activation function and its location to each task, and retains the default when evidence for a change is insufficient. In an evaluation supported by 1,225 Informer training runs across 17 dataset variants and 51 forecasting tasks, the policy reduces mean squared error by 7.86\% on average relative to the conventional fixed-activation baseline and improves 72.55\% of tasks. These findings establish activation choice and placement as practical opportunities for improving existing forecasting models and show how accumulated experimental evidence can guide configuration decisions without repeating an exhaustive search for every task.}
\let\linenumbers\nolinenumbers
\begin{document}
\nolinenumbers
\crefname{algorithm}{Algorithm}{Algorithms}
\Crefname{algorithm}{Algorithm}{Algorithms}
\section{Introduction}

This study asks how forecasting accuracy can be improved by choosing an activation function and its location to suit the task. We show that the best choice changes across datasets and prediction horizons, and turn this variation into a practical selection policy. The aim is to make better use of an existing forecasting model by learning from completed experiments, rather than redesigning its architecture for each new task.\footnote{\raggedright Code repository: \artifacturl{}.\par}

The public Informer implementation uses GELU as its default activation \citep{informerimplementation}; in our controlled experiments, this setting defines the reference configuration. The choice is consequential because every feed-forward block repeatedly transforms temporal representations through that nonlinearity, controlling saturation, sign preservation, and amplification. Placement introduces a second axis: changing encoder FFNs, decoder FFNs, or the output head need not produce the same behavior. The resulting question is broader than whether one activation beats another: \emph{is activation configuration a fixed architectural constant, or a task-dependent decision that should be calibrated?}

We answer the first part with a matched Informer study in which architecture, data split, optimizer, and training budget are fixed while activation type and placement vary. The complete grid contains 972 seed-runs over 36 dataset--horizon cells. A non-GELU configuration is best in 34 cells, and the identity of that configuration changes across horizons and datasets. Bounded hidden activations are strong global candidates, whereas encoder-, decoder-, and output-side interventions win in distinct regimes. Activation is therefore neither a harmless default nor a one-time global substitution; it defines a structured configuration portfolio.

This portfolio view suggests a policy stack with increasing specificity. The conventional GELU default requires no calibration. Our first policy layer, the \emph{calibration-fold single-best configuration} (CF-SBC), selects one activation configuration from previously evaluated datasets and deploys it globally. Our second layer uses input-derived statistical and time--frequency factors---including volatility, jumps, tail behavior, spectral structure, and horizon pressure---to conditionally override CF-SBC. A deterministic verifier accepts an override only when its calibration support, precision, and expected uplift pass fixed thresholds; low-evidence tasks retain CF-SBC. Thus the commonly named ``train-static baseline'' is not external to the proposed approach: it is the stable base policy on which conditional routing operates.

Across 17 Informer dataset variants and 51 dataset--horizon tasks, this policy stack reduces mean MSE by 7.86\% relative to GELU and is positive in 72.55\% of cells. CF-SBC already captures 6.55\% by replacing an uncalibrated default with an evidence-selected global action. Factor-conditioned refinement contributes a further 0.89\% and reduces mean oracle regret from 6.10\% to 4.82\%, a 20.9\% relative reduction. At routing time the policy requires neither target-label access nor backbone retraining, and its frozen templates are executed without an LLM call.

The main contributions of this paper are summarized as follows.
\begin{itemize}
    \item We expose activation type and placement as a structured LTSF configuration portfolio. A 972-run complete grid provides matched evidence that its optimum is dataset- and horizon-dependent even when the remaining forecasting protocol is unchanged.
    \item We introduce a calibrated activation policy that unifies global selection and conditional routing: CF-SBC supplies the robust base action, while factor rules provide supported overrides with explicit guards and abstention.
    \item We evaluate the policy over 1,225 Informer seed-runs and 51 tasks, quantify how each policy layer closes the oracle gap, and compare it with deterministic CART-, RIPPER-, and SATzilla-style selectors under the same leave-dataset protocol.
\end{itemize}

\section{Related Work}

Prior research provides three routes to choosing nonlinear transformations: designing activation functions, learning or searching for task-specific activations, and selecting among existing model configurations. We review these routes to explain where activation selection has already been studied and how our work uses prior forecasting experiments to make interpretable configuration choices for new tasks.

\subsection{Non-linear activations and tactical placement}
ReLU \citep{nair2010relu} and GELU \citep{hendrycks2016gelu} define widely used nonlinear transformations. Their original papers establish the functions, while the GELU default used in this study follows the public Informer implementation \citep{informerimplementation}. Prior work has explored learnable, searched, and periodic activations, including parametric rectifiers \citep{he2015prelu}, automated discovery of Swish-like functions \citep{ramachandran2017searching}, and sinusoidal representation networks for high-frequency implicit signals \citep{sitzmann2020siren}. These studies establish that activation form matters. Dynamic ReLU generates activation parameters from the input, and meta-ACON learns the degree of nonlinearity \citep{chen2020dynamic,ma2021acon}. In long-term forecasting, TC-KAN conditions activation coefficients on temporal position \citep{shen2026tckan}; this position denotes a sequence index rather than an encoder, decoder, or output insertion point. These approaches adapt the transformation itself, whereas our policy selects among explicit configurations using evidence from previously evaluated forecasting tasks. Our work therefore moves from global replacement to conditional selection, and it treats both the activation regime and the insertion position as explicit modeling variables.

\subsection{Dynamic routing and conditional computation}
Our framework is related in spirit to conditional computation, where computation is allocated selectively rather than applied uniformly \citep{bengio2013conditional}. Mixture-of-Experts methods formalize this idea through expert modules and learned gates, from adaptive local experts \citep{jacobs1991adaptive} to sparsely gated MoE layers \citep{shazeer2017outrageously} and Switch Transformers \citep{fedus2022switch}. These systems are powerful, but their routing decisions are learned as part of the model and typically introduce extra training objectives, load-balancing constraints, and optimization failure modes. For forecasting tasks with limited regimes and noisy temporal factors, such trainable gates can be heavier than the activation-selection problem requires. We instead use a deterministic factor router over raw statistical and time-frequency indicators. The key difference is not merely lower cost: the router can abstain when the evidence for intervention is weak, falling back to the default activation instead of forcing every input through a learned routing decision.

\subsection{Algorithm selection and performance portfolios}
The activation portfolio is also an instance of the classical algorithm-selection problem: map measurable instance features to the candidate with the best expected performance \citep{rice1976algorithm}. Portfolio systems distinguish a single-best solver from an oracle or virtual-best solver and ask how much of that gap a selector closes \citep{bischl2016aslib}. SATzilla (2008) uses empirical hardness models to predict solver runtime or competition score before selecting a solver \citep{xu2008satzilla}; CART and RIPPER provide auditable tree and rule-list alternatives \citep{breiman1984cart,cohen1995ripper}. We adapt this perspective to LTSF tasks rather than individual solver instances. CF-SBC plays the role of the single-best policy, the per-cell activation optimum defines the oracle, and factor routing is evaluated by forecasting gain and oracle regret rather than exact-label accuracy alone.

\subsection{Architectural optimization and orthogonality in LTSF}
LTSF progress has been dominated by architecture-centered innovations, including efficient attention in Informer \citep{zhou2021informer}, decomposition and auto-correlation in Autoformer \citep{wu2021autoformer}, frequency-enhanced decomposition in FEDformer \citep{zhou2022fedformer}, patch-based representations in PatchTST \citep{nie2023patchtst}, inverted variate-token modeling in iTransformer \citep{liu2024itransformer}, and linear baselines that question whether Transformer complexity is always necessary \citep{zeng2023transformers}. Automated architecture search extends this line by treating activations as one option inside a much larger combinatorial or differentiable space; AutoFormer-TS, for example, searches over attention mechanisms, activation functions, and encoding operations \citep{zhang2025autoformerts}. AutoFormer-TS permits different activations in different blocks and examines activation-space ablations and cross-task architecture transfer. Its search and training stages optimize an architecture for a task; our method instead selects a configuration in a fixed architecture from a previously measured portfolio. Configuration search remains a computational decision as well as a modeling decision \citep{bergstra2012random}. Our policy instead reuses completed calibration evidence, does not update backbone weights during configuration selection, and selects an activation action with negligible routing-time cost.

\section{Empirical Motivation: Activation Conditionality}

The experiments in this section show why activation configuration should depend on the forecasting task. They compare activation functions and insertion locations under matched training conditions, revealing changes in the best configuration across datasets, prediction horizons, and forecasting architectures.

\subsection{GELU is not consistently optimal}

The complete Informer activation grid is the main evidence for activation conditionality. It covers 36 dataset-horizon cells, nine activation configurations, and three seeds per configuration, for 972 matched seed-runs. \Cref{tab:phase2_placement} reports representative individual-seed MSE and MAE comparisons; aggregate findings use the complete grid. Non-GELU configurations are best in \(34/36\) cells: all 20 ETT/ETTm cells and 14 of 16 Solar cells prefer a non-GELU activation, while Solar3 at horizons 24 and 96 remains GELU-best.

Under a matched LTSF protocol, activation regime is therefore an exposed design variable whose best choice depends on the dataset--horizon task.

\subsection{No single activation dominates}

\begin{figure}[!htbp]
\centering
\includegraphics[width=\linewidth]{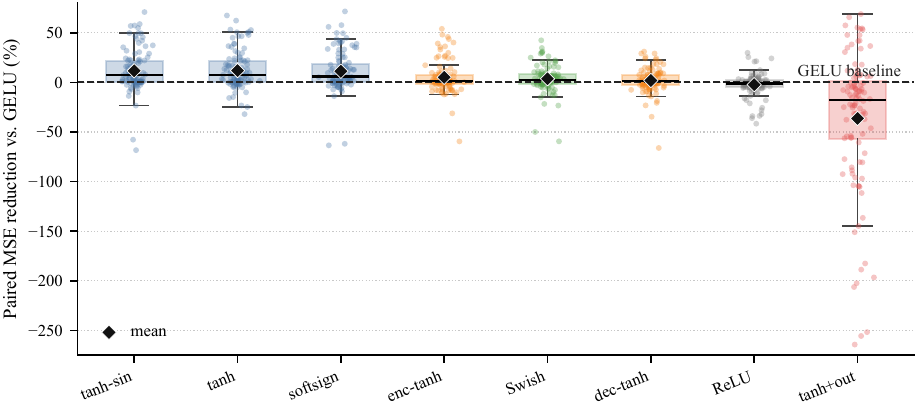}
\caption{Raw paired gain distribution for non-GELU activation configurations. Dots are matched seed-run comparisons against \texttt{gelu\_all}; each configuration has 108 raw points and total comparisons are 864; boxes show median/IQR, whiskers use 1.5 IQR, and diamonds mark means}
\label{fig:phase4_static_ranking}
\end{figure}

\Cref{fig:phase4_static_ranking} replaces an aggregate ranking with the raw paired seed-run distribution behind the complete static-regime comparisons. Each point is a matched comparison against the same dataset--horizon--seed \texttt{gelu\_all} run. Bounded tanh-family activations have positive medians and means, making them effective candidates for the global policy layer. The dispersion also motivates a conditional layer: GELU remains preferable in selected cells, and output-side \texttt{tanh} combines large task-specific gains with a broad lower tail.

This distribution motivates the two-level policy used in this paper. Global calibration captures the strong average behavior of bounded activations; factor-conditioned overrides preserve the ability to select placement actions or retain a safer base configuration when task evidence differs.

\subsection{Placement matters}

Changing the activation family is only one axis. \Cref{tab:phase2_placement} illustrates matched placement and baseline-control results. Hidden tanh helps ETTh1 cells, output \texttt{tanh} gives the best result on selected ETTh2 cells, encoder-side tanh is strongest on selected Solar cells, and output \texttt{tanh} is harmful on Solar96. These results indicate that activation placement can change the model response as much as activation regime.

Table~\ref{tab:phase2_placement} presents 36 illustrative configurations across 12 forecasting cases, spanning short and long horizons, hidden and output placements, a GELU-best Solar case, held-out datasets, and three forecasting architectures. Each case retains its matched GELU control and two activation alternatives, with every available seed reported. Appendix~\ref{app:complete_results} (Table~\ref{tab:all_seed_results}) reports all 543 configurations and 1,471 unique seed-runs supporting the manuscript (1,225 Informer, 198 PatchTST, and 48 iTransformer). The accompanying CSV retains the full-precision values. All aggregate comparisons use that complete evidence; the compact table illustrates the mechanisms behind the results.

\begin{table}[p]\centering\footnotesize
\setlength{\tabcolsep}{3pt}
\renewcommand{\arraystretch}{1.05}
\caption{Representative activation comparisons: individual-seed MSE and MAE (lower is better). Cases illustrate horizon, placement, baseline retention, and backbone effects. Seeds S1--S3 are 2024--2026; a dash denotes an unobserved run. Complete results are reported in Appendix~\ref{app:complete_results}, Table~\ref{tab:all_seed_results}.}\label{tab:phase2_placement}
\begin{tabular}{@{}lllr rrrrrr@{}}
\toprule Dataset & $L/H$ & Config. & & \multicolumn{2}{c}{S1} & \multicolumn{2}{c}{S2} & \multicolumn{2}{c}{S3}\\
& & & & MSE & MAE & MSE & MAE & MSE & MAE\\\midrule

\multicolumn{10}{l}{\textbf{Informer}}\\*
ETTh1 & 48/24 & GELU &  & 0.5368 & 0.5253 & 0.6383 & 0.5824 & 0.5189 & 0.5164\\*
ETTh1 & 48/24 & ReLU &  & 0.5449 & 0.5218 & 0.6683 & 0.5963 & 0.5355 & 0.5223\\*
ETTh1 & 48/24 & tanh &  & 0.4975 & 0.4986 & 0.5593 & 0.5376 & 0.5423 & 0.5337\\
\addlinespace[2pt]
ETTh1 & 336/720 & GELU &  & 1.3769 & 0.9390 & 1.4686 & 0.9883 & 1.3924 & 0.9710\\*
ETTh1 & 336/720 & tanh &  & 1.3994 & 0.9709 & 1.4758 & 1.0160 & 1.5320 & 1.0352\\*
ETTh1 & 336/720 & tanh+out &  & 1.2958 & 0.9113 & 1.3555 & 0.9542 & 1.3322 & 0.9460\\
\addlinespace[2pt]
ETTh2 & 336/168 & GELU &  & 4.8943 & 1.8629 & 4.1212 & 1.7539 & 6.4105 & 2.1213\\*
ETTh2 & 336/168 & tanh &  & 2.5559 & 1.2781 & 3.1508 & 1.3658 & 3.5263 & 1.4566\\*
ETTh2 & 336/168 & tanh+out &  & 3.0216 & 1.3905 & 2.6107 & 1.3376 & 3.3568 & 1.4740\\
\addlinespace[2pt]
ETTm1 & 672/288 & GELU &  & 1.1021 & 0.8160 & 0.8389 & 0.6918 & 0.7159 & 0.6367\\*
ETTm1 & 672/288 & tanh &  & 0.8344 & 0.7275 & 0.8193 & 0.7148 & 0.7521 & 0.6734\\*
ETTm1 & 672/288 & tanh-sin &  & 0.8589 & 0.7408 & 0.7623 & 0.6875 & 0.7882 & 0.6909\\
\addlinespace[2pt]
Solar4 & 96/96 & GELU &  & 0.2978 & 0.3139 & 0.2990 & 0.3146 & 0.2920 & 0.3132\\*
Solar4 & 96/96 & tanh &  & 0.2938 & 0.3203 & 0.2947 & 0.3187 & 0.2897 & 0.3164\\*
Solar4 & 96/96 & enc-tanh &  & 0.2899 & 0.3163 & 0.2861 & 0.3166 & 0.2953 & 0.3196\\
\addlinespace[2pt]
Solar1 & 96/96 & GELU &  & 0.2250 & 0.2225 & 0.2293 & 0.2264 & 0.2355 & 0.2296\\*
Solar1 & 96/96 & tanh &  & 0.2171 & 0.2196 & 0.2331 & 0.2319 & 0.2241 & 0.2304\\*
Solar1 & 96/96 & tanh+out &  & 0.4057 & 0.3185 & 0.4065 & 0.3187 & 0.4035 & 0.3167\\
\addlinespace[2pt]
Solar3 & 96/96 & GELU &  & 0.3415 & 0.3560 & 0.3556 & 0.3667 & 0.3499 & 0.3622\\*
Solar3 & 96/96 & tanh &  & 0.3952 & 0.3989 & 0.3336 & 0.3781 & 0.3866 & 0.4029\\*
Solar3 & 96/96 & enc-tanh &  & 0.3819 & 0.3888 & 0.3512 & 0.3713 & 0.3515 & 0.3719\\
\addlinespace[2pt]
Electricity & 96/96 & GELU &  & 0.2817 & 0.3762 & 0.2730 & 0.3685 & 0.2782 & 0.3718\\*
Electricity & 96/96 & tanh &  & 0.2928 & 0.3869 & -- & -- & -- & --\\*
Electricity & 96/96 & dec-tanh &  & 0.2773 & 0.3747 & 0.2668 & 0.3652 & 0.2783 & 0.3733\\
\addlinespace[2pt]
Traffic & 96/192 & GELU &  & 0.7253 & 0.3946 & 0.7674 & 0.4143 & 0.7577 & 0.4048\\*
Traffic & 96/192 & tanh &  & 0.7545 & 0.4077 & -- & -- & -- & --\\*
Traffic & 96/192 & dec-tanh &  & 0.7353 & 0.4005 & -- & -- & -- & --\\
\addlinespace[2pt]
Traffic & 96/336 & GELU &  & 0.7667 & 0.4218 & 0.8186 & 0.4426 & 0.7525 & 0.4112\\*
Traffic & 96/336 & tanh &  & 0.7384 & 0.4063 & -- & -- & -- & --\\*
Traffic & 96/336 & dec-tanh &  & 0.7564 & 0.4162 & -- & -- & -- & --\\
\addlinespace[2pt]
\multicolumn{10}{l}{\textbf{PatchTST}}\\*
ETTh2 & 336/336 & GELU &  & 0.3287 & 0.3847 & 0.3288 & 0.3851 & 0.3278 & 0.3833\\*
ETTh2 & 336/336 & tanh &  & 0.3328 & 0.3817 & 0.3349 & 0.3830 & 0.3319 & 0.3814\\*
ETTh2 & 336/336 & tanh+out &  & 0.3395 & 0.3856 & 0.3443 & 0.3886 & 0.3384 & 0.3857\\
\addlinespace[2pt]
\multicolumn{10}{l}{\textbf{iTransformer}}\\*
ETTm2 & 96/720 & GELU &  & 0.4141 & 0.4065 & -- & -- & -- & --\\*
ETTm2 & 96/720 & tanh &  & 0.4145 & 0.4059 & -- & -- & -- & --\\*
ETTm2 & 96/720 & tanh+out &  & 0.4085 & 0.4025 & -- & -- & -- & --\\
\addlinespace[2pt]
\bottomrule
\end{tabular}
\par\smallskip
\noindent\textbf{Configuration key.} Plain labels denote hidden-FFN activations; enc-tanh and dec-tanh change only the respective Informer stack. The suffix +out adds tanh after the output projection. The tanh-sin perturbation amplitude is 0.01. $L/H$ denotes input/prediction length.
\par\end{table}

We therefore treat placement as part of the action space rather than as an implementation detail. The policy distinguishes static all-layer regimes from encoder-side, decoder-side, and output-side interventions.

\subsection{Architecture dependence}

Our transfer probes show that the configuration axes persist but must be instantiated per architecture. In our PatchTST experiment, non-GELU configurations are best in 5 of 12 three-seed regime cells, whereas the matched output-tanh probe favors the linear head in all 54 comparisons. In our iTransformer experiment, the FFN-only regime effect is small, while placement interventions are positive in all three one-seed probe cells. These results identify an architecture-specific portfolio rather than a universal ordering: the main policy is evaluated on Informer, and PatchTST/iTransformer serve as controlled evidence that action definitions must respect backbone topology.

\section{Method}

Our method chooses which activation function to use and where to apply it for a new forecasting task. It first identifies a reliable default from previous experiments, then checks whether the new task's characteristics support a better alternative. If the evidence supports a change, it selects that alternative; otherwise, it keeps the default. The following sections define these choices and present the complete procedure as pseudocode.

\subsection{Problem formulation and action space}

Given a multivariate input window \(X_{t-L+1:t}\in\mathbb{R}^{L\times C}\), the forecasting model predicts the next \(H\) target values \(Y_{t+1:t+H}\). For activation configuration \(a\), dataset variant \(d\), horizon \(H\), and seed \(s\), let \(m(a,d,H,s)\) denote MSE. Improvements over the default are reported as paired relative reductions:
\begin{equation}
    G(a,d,H,s)
    =
    \frac{m(\texttt{gelu\_all},d,H,s)-m(a,d,H,s)}
         {m(\texttt{gelu\_all},d,H,s)}.
\end{equation}
Positive \(G\) means that the action improves over the same-seed GELU run under the same dataset-horizon protocol.

The policy uses the nine auditable configurations in \Cref{tab:activation_action_space}. The suffix \texttt{all} refers specifically to every hidden FFN in both the encoder and decoder; it does not change attention projections, temporal embeddings, convolutional distillation, normalization, or the final projection. Static-regime actions change both hidden FFN stacks. Placement actions change one stack or add tanh after the linear output projection. The output action is kept separate because it bounds the forecast itself rather than only transforming a hidden representation.

\begin{table*}[t]
\centering
\caption{Orthogonal definition of the nine Informer activation configurations. ``Encoder'' and ``decoder'' refer to the nonlinearities between the two convolutions in every corresponding FFN block. Attention projections, temporal embeddings, normalization, and all other modules are unchanged. Output activation is applied after the final linear projection}
\label{tab:activation_action_space}
\small
\setlength{\tabcolsep}{4pt}

\begin{tabularx}{\linewidth}{@{}lXXl@{}}
\toprule
Configuration & Encoder FFNs & Decoder FFNs & Output head \\
\midrule
\texttt{gelu\_all} & GELU & GELU & linear \\
\texttt{relu\_all} & ReLU & ReLU & linear \\
\texttt{swish\_all} & SiLU/Swish & SiLU/Swish & linear \\
\texttt{tanh\_all} & tanh & tanh & linear \\
\texttt{softsign\_all} & softsign & softsign & linear \\
\texttt{tanh\_sin001\_all} & $\tanh(x)+0.01\sin(x)$ & $\tanh(x)+0.01\sin(x)$ & linear \\
\texttt{enc\_tanh\_dec\_gelu} & tanh & GELU & linear \\
\texttt{enc\_gelu\_dec\_tanh} & GELU & tanh & linear \\
\texttt{tanh\_all\_outtanh} & tanh & tanh & tanh after projection \\
\bottomrule
\end{tabularx}
\end{table*}

\subsection{Input-only factors and factor buckets}

For each dataset-horizon task, factors are computed only from the input window and task metadata. They include normalized volatility \(\sigma(X)/(|\mu_{\mathrm{ref}}|+\epsilon)\), realized volatility, volatility shock relative to a previous reference window, jump intensity, tail ratio, trend consistency, reversal rate, mean-crossing rate, lag-1 autocorrelation of differences, spectral entropy, high-frequency energy share, dominant-period ratios, and the horizon ratio \(H/L\). These factors describe the task before observing target-cell activation results. Target-cell MSE is never used as a router input.

\subsection{Cost-aware hierarchical decision structure}

The policy optimizes forecasting cost rather than exact winner labels. Exact-best classification assigns the same error to a near-tie and a catastrophic mismatch, although their deployment costs differ sharply. We instead preserve the continuous MSE surface and organize selection hierarchically: establish the strongest calibration-fold action, test whether input factors justify an override, select a coarse action family, and map that family to an exact configuration using calibration performance. When an override lacks support, the policy retains the calibrated base action.

\subsection{Hierarchical calibrated router}

For a held-out target dataset variant \(d^\star\), the calibration fold is
\begin{equation}
    \mathcal{T}(d^\star)=\{(d,H): d\neq d^\star\}.
\end{equation}
The base policy is the calibration-fold single-best configuration
\begin{equation}
    a_{\mathrm{CF\text{-}SBC}}(d^\star)
    = \arg\min_{a\in\mathcal{A}}
      \frac{1}{|\mathcal{T}(d^\star)|}
      \sum_{(d,H)\in\mathcal{T}(d^\star)} \overline{m}(a,d,H),
\end{equation}
where \(\overline{m}\) averages the available matched seeds within a cell. CF-SBC converts prior activation evidence into a deployable global configuration without accessing the target dataset's activation outcomes.

For each candidate action \(a\), the calibration fold then estimates pairwise uplift against \texttt{gelu\_all} and CF-SBC. A conditional rule is accepted only when its support, precision, and mean uplift exceed fixed thresholds. In the balanced setting used for the main result, the minimum objective gain is \(0.002\) and the minimum precision is \(0.65\). Low-confidence cases retain CF-SBC; a GELU-fallback variant is included as an ablation.

For a rule \(r\) and action \(a\), let \(F_r(d,H)\in\{0,1\}\) indicate that the rule fires on a calibration cell. Its support is \(\sum_{\mathcal{T}}F_r\), precision is the fraction of fired cells with uplift above margin, and mean uplift is the average fired-cell uplift. Candidate rules are ranked by objective, precision, and fired uplift, with action-family selection performed before exact-config selection.

\subsubsection{Factor-bucket implementation}
The factor-bucket implementation is inspired by factor analysis in quantitative finance: a factor is useful when sorting cells by that factor creates a stable spread in a downstream response. We replace financial return with activation uplift. For a dataset-horizon cell \(\gamma=(d,H)\), action \(a\), input factor \(f_j\), and calibration-fold quantiles \(q^{\mathrm{train}}_{j,m}\), the \(m\)-th bucket and its action response are
\begin{subequations}
\renewcommand{\theHequation}{\theparentequation\alph{equation}}
\begin{align}
    B_{j,m}
    &= \{\gamma: q^{\mathrm{train}}_{j,m-1} \le f_j(\gamma) < q^{\mathrm{train}}_{j,m}\}, \label{eq:factor_bucket_set}\\
    u(\gamma,a)
    &= \frac{\mathrm{MSE}(\gamma,\texttt{gelu\_all})-\mathrm{MSE}(\gamma,a)}
            {\mathrm{MSE}(\gamma,\texttt{gelu\_all})+\epsilon}, \label{eq:factor_bucket_uplift}\\
    b_{j,m}(a)
    &= \frac{1}{|B_{j,m}|}\sum_{\gamma\in B_{j,m}} u(\gamma,a). \label{eq:factor_bucket_response}
\end{align}
\end{subequations}
Buckets are used for interpretable rule induction rather than as a high-dimensional classifier: repeated positive or negative spreads become candidate templates, and low-support buckets trigger abstention or rejection by the deterministic verifier.

Among the larger factor pool, we use five representative factors in the paper because repeated heuristic bucket experiments found that they cover the main observed mechanisms: amplitude variation, shocks, tails, spectral concentration, and horizon pressure. Let \(x_{t,r}\) denote channel \(r\) at input time \(t\), \(z_{t,r}=(x_{t,r}-\mu_r)/(\sigma_r+\epsilon)\), \(\Delta z_{t,r}=z_{t,r}-z_{t-1,r}\), and
\(p_{k,r}=|\mathrm{FFT}(z_{\cdot,r})_k|^2/\sum_{\ell=1}^{K}|\mathrm{FFT}(z_{\cdot,r})_\ell|^2\). The representative factors are
\begin{subequations}
\renewcommand{\theHequation}{\theparentequation\alph{equation}}
\begin{align}
    f_{\mathrm{vol}}
    &= \frac{1}{C}\sum_{r=1}^{C}
       \sqrt{\frac{1}{L-1}\sum_{t=2}^{L}(\Delta z_{t,r})^2}, \label{eq:representative_factor_vol}\\
    f_{\mathrm{jump}}
    &= \frac{1}{C(L-1)}\sum_{r=1}^{C}\sum_{t=2}^{L}
       \mathbb{I}\{|\Delta z_{t,r}| > 3\,\operatorname{MAD}(\Delta z_{\cdot,r})\}, \label{eq:representative_factor_jump}\\
    f_{\mathrm{tail}}
    &= \frac{1}{C}\sum_{r=1}^{C}
       \frac{Q_{0.95}(|z_{\cdot,r}|)}{Q_{0.50}(|z_{\cdot,r}|)+\epsilon}, \label{eq:representative_factor_tail}\\
    f_{\mathrm{spec}}
    &= \frac{1}{C}\sum_{r=1}^{C}
       \left[-\frac{\sum_{k=1}^{K}p_{k,r}\log(p_{k,r}+\epsilon)}{\log K}\right], \label{eq:representative_factor_spec}\\
    f_H
    &= H/L . \label{eq:representative_factor_horizon}
\end{align}
\end{subequations}
These five factors form a compact, interpretable state description: volatility and jumps mark unstable inputs, tails identify conditions in which output bounding is risky, spectral entropy separates concentrated from dispersed frequency structure, and \(H/L\) captures horizon-dependent action switches.

\subsection{Frozen template induction and deterministic execution}

\Cref{alg:policy} presents the complete workflow: propose candidate rules, verify them using prior tasks, and select a configuration for the new task. The protocol serializes factor--action response summaries into a hash-recorded induction packet containing input-derived factors, task metadata, action names and families, and continuous calibration-side uplift. A normalized authoring prompt asks either an LLM or a human to return JSON hypotheses only: a template identifier and type, an action and family, a conjunction or disjunction of one to five factor conditions, and a short mechanism or failure hypothesis. The authoring tool cannot mark a rule as accepted and never emits a test-time decision. A JSON schema checks the action and factor vocabulary, required fields, condition count, directions, and quantile-versus-threshold format before the candidate set is frozen.

Table~\ref{tab:router_template_examples} illustrates the frozen rule conditions, actions, calibration support, and extrapolation flags. The offline candidate-authoring step used OpenAI Codex with model identifier \texttt{gpt-5.5} and reasoning effort \texttt{xhigh}. Appendix~\ref{app:template_induction_protocol} provides the recovered historical task prompt for reference, its English translation, the complete reusable prompt, and execution instructions. Recovered creation and revision records reproduce the evaluated JSON byte for byte. The policy is evaluated by loading that frozen JSON and its deterministic verifier. Three templates are explicitly marked as extrapolation hypotheses, so \Cref{fig:policy_evidence}c reports both the all-cell result and the scope that removes those rows.

For each held-out variant \(d^\star\), the verifier receives its input factors, the hash-identified template set, and \(\mathcal{T}(d^\star)\). Quantile thresholds, rule support, precision, mean uplift, CF-SBC, and within-family configuration rankings are computed after removing all outcome rows from \(d^\star\). Guard rules have priority. A matched conditional rule can override CF-SBC; if none qualifies, CF-SBC is returned. Target-cell MSE is unlocked only after the action has been emitted, for gain and regret reporting.

\begin{algorithm}[t]
\caption{Activation configuration selection for a new task}\label{alg:policy}
\begin{algorithmic}[1]
\Require Prior task results $D$, candidate configurations $\mathcal A$, new task $d^\star$ and its input observations $X^\star$
\Ensure Selected activation configuration $a^\star$ and decision record
\State \textbf{Offline preparation}
\State Summarize how input characteristics relate to activation performance
\State Use offline language-model assistance to propose candidate rules $\mathcal R$
\State Check the rule schema and freeze the candidate set
\State \textbf{Calibration for the target task}
\State $\mathcal T\gets D$ excluding all results from dataset $d^\star$
\State $a_0\gets$ configuration with lowest average calibration MSE
\For{each candidate rule $r\in\mathcal R$}
  \State Estimate its thresholds, support, precision, and uplift on $\mathcal T$
  \State Retain $r$ if it meets all acceptance thresholds
\EndFor
\State \textbf{Configuration selection}
\State Compute input factors $x^\star$ from $X^\star$ and task metadata
\If{an accepted guard rule matches $x^\star$}
  \State $a^\star\gets$ the guard's designated configuration
\ElsIf{accepted conditional rules match $x^\star$}
  \State Rank the matched families by calibration objective, precision, and uplift
  \State $a^\star\gets$ lowest-calibration-MSE configuration in the winning family
\Else
  \State $a^\star\gets a_0$
\EndIf
\State \Return $a^\star$ and the matched rules and calibration scores
\Statex Target outcomes are used subsequently for evaluation, not for selection.
\end{algorithmic}
\end{algorithm}

\begin{table*}[t]
\centering
\caption{Examples of frozen offline-assisted templates}
\label{tab:router_template_examples}
\small
\setlength{\tabcolsep}{3pt}
\renewcommand{\arraystretch}{1.15}
\begin{tabularx}{\textwidth}{@{}>{\raggedright\arraybackslash}p{0.23\textwidth}>{\raggedright\arraybackslash}X>{\raggedright\arraybackslash}p{0.18\textwidth}cp{0.07\textwidth}@{}}
\toprule
template & factor condition sketch & action & \shortstack{support /\\precision} & extrap. \\
\midrule
bounded-low-horizon-period & low H/L, high dominant-period-to-horizon ratio & bounded regime & 28 / 0.89 & no \\
tanh-sin-short-bursty & low H/L, high time-frequency burstiness & tanh-sin & 24 / 0.83 & no \\
decoder-tanh-low-volatility & low input volatility, moderate or long H/L & decoder-side tanh & 15--17 / 0.60--0.73 & no \\
short-high-dim-GELU-guard & Traffic-like short/mid horizon, high autocorr., low HF noise & abstain to GELU & 2 / 1.00 & yes \\
long-horizon-bounded-static & Traffic-like long horizon, high autocorr., low HF noise & bounded static & 1 / 0.65 & yes \\
\bottomrule
\end{tabularx}
\par\smallskip
\begin{minipage}{\textwidth}
\small
1. Templates are executed by the deterministic verifier.\\
2. Full templates are stored as JSON.\\
3. Rows show factor conditions, actions, support, and extrapolation status.
\end{minipage}
\end{table*}

\subsubsection{Reproducibility contract}
The reproducibility materials include the model configuration, recovered task instruction, standalone reference prompt, JSON schema, packet exporter, evaluated templates, rule logs, and deterministic verifier. Appendix~\ref{app:replay} gives executable commands, dependencies, and expected outputs; the complete materials accompany the source package. The project repository address is provided through the \artifactlink{}. The evaluated JSON is anchored by SHA-256 prefix \texttt{67305b22}, and the verifier by \texttt{6a896a76}. The bundled replay reproduces all 408 recorded policy decisions from the frozen factors and costs, including the selected 51-task comparison. Candidate regeneration follows the reference prompt and yields a new artifact to validate and freeze.

\subsection{Policy references and oracle}

The evaluation separates the three policy levels. \texttt{gelu\_all} is the conventional uncalibrated default. CF-SBC is the global calibration layer and policy fallback. The complete factor policy adds conditional overrides. The held-out oracle is the best available configuration in the target cell, including GELU, and is used only as an evaluation upper bound. Primary metrics are total gain over GELU, incremental gain over CF-SBC, positive-cell rate, oracle regret, and abstention rate. Exact-configuration and action-family alignment are diagnostics because performance-equivalent actions can have different hard labels. We additionally compare CART classification, CART cost regression, RIPPER pairwise rules, and SATzilla-style per-action cost models under the same outer leave-dataset protocol.
\FloatBarrier

\section{Experiments}

The experiments test whether task-specific activation selection improves forecasting accuracy and which parts of the method produce that improvement. We report individual training runs, compare the complete policy with fixed choices and alternative selectors, and examine how its behavior changes across datasets, prediction horizons, random seeds, and forecasting architectures.

\subsection{Experimental design and leakage control}

The experiments ask whether a calibrated policy can choose an activation regime or placement before the target cell's activation outcomes are available, while keeping the forecasting backbone and training protocol fixed. \Cref{tab:router_evidence_coverage} summarizes 1,225 unique Informer seed-runs over 17 dataset variants and 51 dataset--horizon cells: a complete three-seed activation grid, a three-seed extension, a one-seed Electricity/Traffic stress block, and targeted three-seed confirmation of selected decisions. The 51 cells, rather than the individual seeds, are the independent selector examples; seeds estimate each cell--action cost. All aggregation uses canonical dataset, horizon, configuration, and seed keys. The 51-task selector comparison retains the original stress-test measurements; the later confirmation seeds are evaluated separately and are included in Appendix~\ref{app:complete_results} (Table~\ref{tab:all_seed_results}), with representative cases in Table~\ref{tab:phase2_placement}.

\begin{table*}[t]
\centering
\caption{Evidence scope for calibrated activation-policy evaluation}
\label{tab:router_evidence_coverage}
\small
\begin{tabular*}{\linewidth}{@{\extracolsep{\fill}}lrrrrr@{}}
\toprule
Block & variants & cells & configs & seeds & runs \\
\midrule
Complete activation grid & 12 & 36 & 9 & 3 & 972/972 \\
Oracle extension & 3 & 9 & 7 & 3 & 189/189 \\
Held-out stress test & 2 & 6 & 7 & 1 & 42/42 \\
Targeted router check & 2 & 5 & targeted & 3 & 33/33 \\
\midrule
Unique Informer total & 17 & 51 & -- & -- & 1,225 \\
\bottomrule
\end{tabular*}
\par\smallskip
\begin{minipage}{\textwidth}
\footnotesize
1. Main-grid and oracle-extension rows use complete three-seed matrices.\\
2. Electricity/Traffic is a one-seed held-out stress test.\\
3. Targeted checks combine 11 existing runs with 22 new confirmation runs.\\
4. After removing that overlap, the unique Informer total is 1,225 seed-runs.
\end{minipage}
\end{table*}

The Informer protocol uses ProbSparse attention, time-feature embeddings, \(d_{\mathrm{model}}=512\), \(d_{\mathrm{ff}}=2048\), eight heads, dropout \(0.05\), Adam with initial learning rate \(10^{-4}\), patience \(3\), batch size \(32\), and at most six training epochs. ETT originates from the Informer benchmark \citep{zhou2021informer}; Weather uses the Jena station data \citep{weatherjena}; Electricity, Traffic, and Exchange follow the LSTNet/Autoformer benchmark lineage \citep{trindade2015electricity,caltranspems,lai2018lstnet,wu2021autoformer}; and ILI uses aggregate CDC surveillance series \citep{cdcfluview}. Solar1--Solar8 are the eight 15-minute PV station series published by Chen and Xu \citep{chen2022solar,chen2022solardata}, treated as one family in family-collapsed analysis. Appendix~\ref{app:data_protocol} specifies the original sources, exact dataset snapshots, chronological splits, normalization, and backbone-specific training settings. Within each cell, activation regime or placement is the only changed model component.

For each held-out dataset \(d^\star\), CF-SBC, rule support, precision, thresholds, and exact-configuration mapping are computed on \(\mathcal{T}(d^\star)=\{(d,H):d\neq d^\star\}\). All activation outcomes from \(d^\star\) are excluded until the policy has emitted its decision. The result is a leave-dataset evaluation of a frozen calibrated policy; target outcomes serve only for final scoring.

\subsection{Overall policy performance}

\Cref{tab:phase9_router_summary} reports the complete policy. The balanced setting with CF-SBC fallback obtains a \(+7.86\%\) mean MSE reduction over \texttt{gelu\_all}, improves \(72.55\%\) of cells, and leaves \(4.82\%\) mean regret to the held-out oracle. Within that total effect, factor-conditioned overrides add \(+0.89\%\) over CF-SBC and reduce its oracle regret from \(6.10\%\) to \(4.82\%\). The policy abstains to CF-SBC in \(17.65\%\) of cells; action-family alignment is \(80.39\%\), compared with \(33.33\%\) exact-configuration alignment.

\begin{table*}[t]
\centering
\caption{Leave-dataset calibrated activation-policy summary. CF-SBC denotes the calibration-fold single-best configuration}
\label{tab:phase9_router_summary}
\footnotesize
\setlength{\tabcolsep}{2.5pt}
\begin{tabular*}{\textwidth}{@{\extracolsep{\fill}}llrrrrrrr@{}}
\toprule
conf. & fallback & \shortstack{gain vs\\GELU} & \shortstack{incr. vs\\CF-SBC} & regret & \shortstack{pos. vs\\GELU} & \shortstack{exact\\align} & \shortstack{family\\align} & abstain \\
\midrule
balanced & CF-SBC & +7.86\% & +0.89\% & 4.82\% & 72.5\% & 33.3\% & 80.4\% & 17.6\% \\
open & CF-SBC & +7.03\% & -2.05\% & 7.83\% & 76.5\% & 37.3\% & 80.4\% & 7.8\% \\
conservative & CF-SBC & +6.63\% & +0.08\% & 6.02\% & 70.6\% & 23.5\% & 80.4\% & 37.3\% \\
balanced & gelu & +4.52\% & -6.96\% & 12.99\% & 62.7\% & 31.4\% & 74.5\% & 17.6\% \\
open & gelu & +5.17\% & -5.67\% & 11.61\% & 72.5\% & 35.3\% & 78.4\% & 7.8\% \\
conservative & gelu & +4.33\% & -7.15\% & 13.19\% & 51.0\% & 21.6\% & 56.9\% & 37.3\% \\
\bottomrule
\end{tabular*}
\par\smallskip
\begin{minipage}{\textwidth}
\footnotesize
1. Results cover 17 Informer dataset variants and 51 dataset-horizon cells.\\
2. The second gain column isolates the conditional layer beyond the CF-SBC base policy.\\
3. The oracle is the best available configuration in the held-out target cell and may include GELU.
\end{minipage}
\end{table*}

\begin{figure}[!t]
    \centering
    \includegraphics[width=\linewidth]{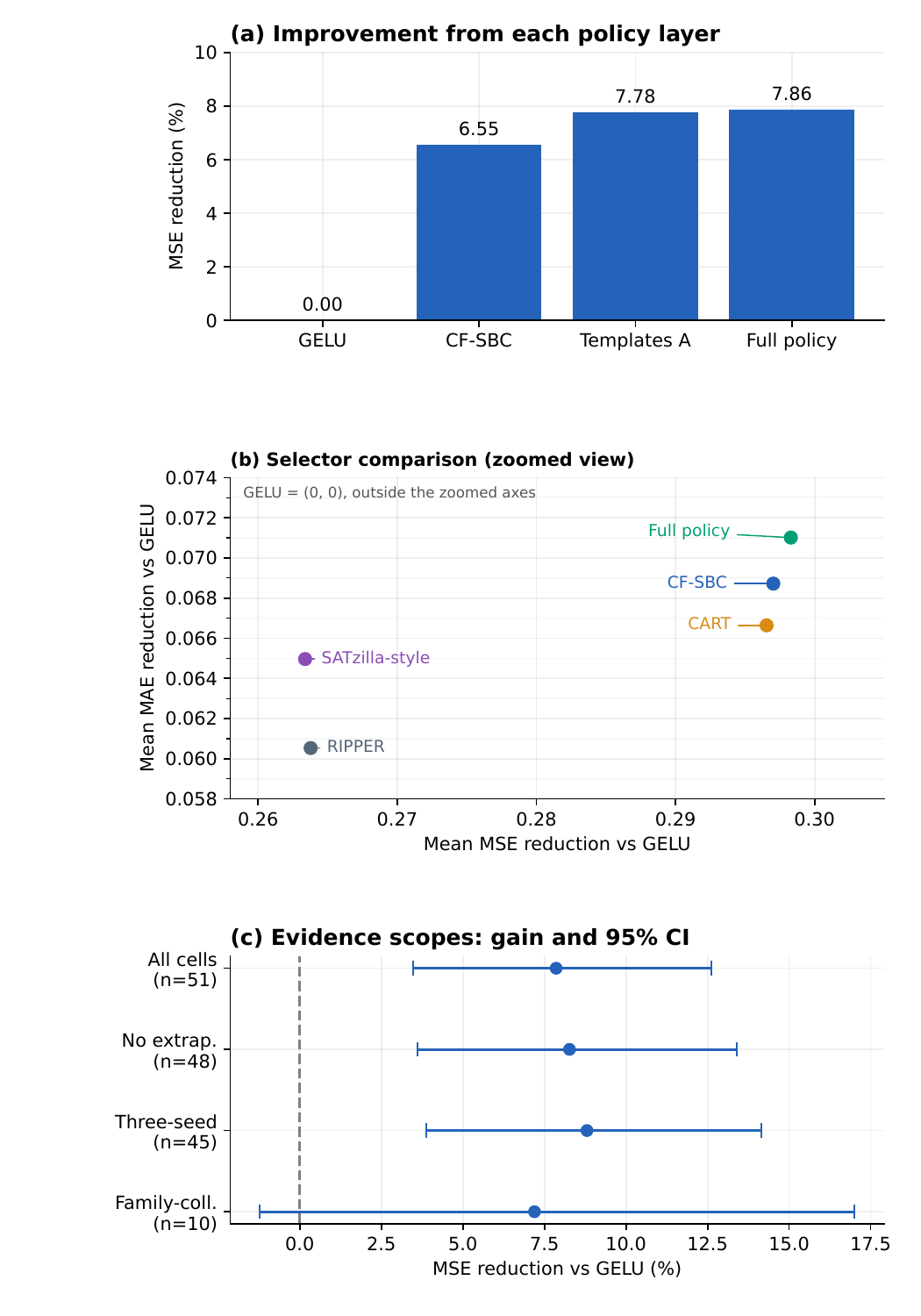}
    \caption{Forecasting improvements from activation selection. (a) Mean relative MSE reduction for each policy layer. (b) Mean absolute MSE and MAE reductions relative to matched GELU results across the same 51 tasks; higher and rightward indicate larger improvements. The axes are zoomed to separate the five selectors; the GELU reference at $(0,0)$ is outside the displayed range. (c) Cell-bootstrap 95\% intervals for relative MSE reduction across four evidence scopes}
    \label{fig:policy_evidence}
\end{figure}

\Cref{fig:policy_evidence} summarizes three complementary views of the same policy evidence: layer-wise construction, comparison with CPU-only classical selectors, and uncertainty across evidence scopes. The same ordering is preserved when analysis is restricted to the 45 three-seed cells: \(+8.80\%\) over GELU and \(+0.68\%\) over CF-SBC. Treating the eight Solar sites as one family yields \(+7.19\%\) over GELU and \(+1.75\%\) over CF-SBC, showing that the aggregate effect is not created by counting Solar sites as independent benchmark families. Appendix~\ref{app:dataset_results} reports the corresponding per-dataset results.

The all-cell bootstrap interval for total gain over GELU is \([+3.47\%,+12.62\%]\), and the cell-level sign test gives \(p<10^{-3}\). The conditional increment over CF-SBC is a positive point estimate with interval \([-0.48\%,+2.62\%]\); the complete policy combines improved mean accuracy with a 20.9\% relative reduction in oracle regret compared with CF-SBC.

\subsection{Policy behavior: alignment and risk control}

\Cref{tab:router_case_studies} shows that the policy realizes its gain through three distinct actions: selecting a bounded regime, selecting a placement intervention, and guarding the base action. Exact bounded-regime hits include ETTh2/24, ETTm1/96, ETTm2/96, ILI/24, selected Solar cells, Weather/96, and Traffic/336. Placement hits include Electricity/96 and /336 and Weather/192 and /336. The \(80.39\%\) family alignment therefore reflects meaningful coarse decisions even when nearby configurations exchange the exact oracle rank.

\begin{table}[!htbp]
\centering
\caption{Representative calibrated-policy action traces}
\label{tab:router_case_studies}
\small
\setlength{\tabcolsep}{2.2pt}
\begin{tabularx}{\linewidth}{@{}llccrr>{\raggedright\arraybackslash}X@{}}
\toprule
trace & cell & policy & oracle & gain & gap & pattern \\
\midrule
bounded & ETTh2/24 & tanh-sin & tanh-sin & +28.73 & 0.00 & bounded short horizon \\
placement & Electricity/96 & dec-tanh & dec-tanh & +1.56 & 0.00 & decoder intervention \\
switch & Traffic/336 & softsign & softsign & +4.00 & 0.00 & horizon-conditioned action \\
family & ETTm1/288 & tanh-sin & tanh & +9.32 & 0.15 & near-equivalent family \\
tail risk & Solar7/24 & tanh-sin & ReLU & -20.53 & 41.86 & rare action family \\
tail risk & Exchange/192 & tanh-sin & dec-tanh & -32.46 & 34.40 & placement mismatch \\
\bottomrule
\end{tabularx}
\par\smallskip
\begin{minipage}{\linewidth}
\footnotesize
1. Cases use leave-dataset calibrated validation.\\
2. Gain/gap are relative MSE percentages against GELU/oracle.\\
3. Tail rows identify where additional family support and abstention are most valuable.
\end{minipage}
\end{table}

The guard mechanism is equally substantive. GELU is oracle in four cells; the short-horizon Traffic guard identifies two of them and retains GELU, while the long-horizon Traffic rule switches to bounded softsign. This horizon-dependent action switch is the behavior that a single global configuration cannot express.

Residual risk is concentrated rather than diffuse. Solar7/24 requires the rare ReLU action, while Exchange/192 requires a decoder-side action that conflicts with the calibration-fold bounded configuration. These tail cases motivate the explicit regret and abstention metrics: they identify where additional family support is more valuable than increasing rule coverage.

\subsection{Policy-layer contribution and selector comparisons}

\Cref{fig:policy_evidence}a builds the policy layer by layer. The GELU default leaves \(19.33\%\) oracle regret. CF-SBC converts calibration evidence into a \(+6.55\%\) gain and cuts regret to \(6.10\%\). With an empty conditional rule set, the policy returns exactly CF-SBC by construction. Adding verified template sets A and B progressively improves total gain and regret; the full policy reaches \(+7.86\%\) and \(4.82\%\), respectively.

The confidence rows in \Cref{tab:phase9_router_summary} isolate the remaining design choices. Replacing CF-SBC fallback with GELU removes much of the globally calibrated gain. The open policy maximizes coverage, the conservative policy maximizes selectivity, and the balanced policy produces the best gain--regret combination. The result is carried by the policy stack---calibrated base, verified overrides, and guards---rather than rule firing alone.

\Cref{fig:policy_evidence}b places the conditional layer against deterministic selector families under the same outer leave-dataset split and grouped inner validation. The proposed policy has the largest mean absolute reductions in both MSE and MAE in this comparison. For metric $E\in\{\mathrm{MSE},\mathrm{MAE}\}$, the plotted improvement is $\Delta E=51^{-1}\sum_{i=1}^{51}(E_{\mathrm{GELU},i}-E_{\mathrm{selector},i})$. The complete policy achieves $\Delta\mathrm{MSE}=0.2983$ and $\Delta\mathrm{MAE}=0.0710$. Each task contributes once after its available seeds are averaged; absolute differences retain each benchmark\'s evaluation scale. By mean relative MSE reduction, the nearest classical alternative is the SATzilla-style per-action cost model at \(+6.67\%\) gain and \(6.30\%\) regret; CART and RIPPER recover useful activation signal but do not match the calibrated hierarchy. All selector fitting and nested validation complete on CPU in under one minute, so the comparison measures selection quality rather than compute budget.

Targeted three-seed confirmation checks the Electricity and Traffic decisions that exercise both placement and guard logic. All five cells are non-degrading versus GELU, three are positive, and their mean effect is \(+0.40\%\) with \(0.31\%\) regret to the available targeted oracle. Electricity/96 and /336 validate decoder-side placement; Traffic/192 validates guarded retention; and Traffic/336 validates the long-horizon switch to a bounded regime.

\subsection{Architecture-specific activation portfolios}

PatchTST and iTransformer provide topology controls because their insertion points differ from Informer. In our PatchTST probe, non-GELU regimes are best in 5 of 12 three-seed cells and deliver a \(+0.38\%\) mean best-action gain; the linear head is preferred in all 54 matched output-placement comparisons. In our iTransformer probe, one of six FFN-regime cells prefers non-GELU, while the mixed configuration portfolio improves all three one-seed placement-probe tasks. The best configurations use output tanh on ETTh2/96 and ETTm2/720; ETTm2/96 instead selects a hidden tanh-sin activation with a linear output head. Thus two tasks support output-placement interventions, and the third supports a hidden-activation change. Together these probes show that activation configuration remains a live variable across backbones, while the admissible portfolio and its calibration must be architecture-aware. Appendix~\ref{app:architecture_results} provides the complete regime and placement summaries.

\FloatBarrier
\section{Conclusion}

Activation configuration is a practical way to improve long-term forecasting within an existing model. This study shows that activation functions and insertion locations suit different tasks, and that prior experimental evidence can guide those choices through a calibrated selection policy.

Our policy stack converts this finding into a low-cost decision procedure: CF-SBC supplies a calibrated global action, and verified factor rules provide supported overrides with abstention back to CF-SBC. Across 51 tasks, it improves mean MSE by 7.86\% over GELU, adds 0.89\% beyond CF-SBC, and reduces oracle regret from 6.10\% to 4.82\%, the strongest observed gain--regret tradeoff among the tested low-cost selectors. This provides an auditable path from fixed defaults to architecture-aware activation policies.

\par
\authorcontributions{Conceptualization, X.S. and B.T.; methodology, X.S. and B.T.; software, X.S. and B.T.; investigation, X.S. and B.T.; formal analysis, X.S., B.T. and K.H.; writing---original draft preparation, X.S. and B.T.; writing---review and editing, X.S., B.T., K.H. and W.J.; project administration, B.T.; supervision, W.J.; resources, W.J.; funding acquisition, W.J. Xu Shijie and Boyan Tang contributed equally to this work and share first authorship. All authors have read and agreed to the submitted version of the manuscript.}
\funding{This research was supported by the Shenzhen Loop Area Institute (Contract No. SLAI2026020007) and the Shenzhen Science and Technology Program (Grant No. C00120260002).}
\institutionalreview{Not applicable. This computational study uses public benchmark time series, including aggregate influenza surveillance statistics, and involves no recruitment, intervention, or identifiable individual-level records.}
\informedconsent{Not applicable. The study does not involve human participants or identifiable personal data.}
\dataavailability{The public dataset sources and access links are listed in Appendix~\ref{app:data_protocol}. Solar data are described by Chen and Xu and archived at \url{https://doi.org/10.6084/m9.figshare.17304221.v4}. The code repository is available through the \artifactlink{}. The accompanying source package contains the full seed-level results, frozen rule artifacts, and a tested decision-replay package. Its version is identified by the file-level SHA-256 manifest; dataset snapshots and training settings are recorded in \texttt{artifacts/data\_protocol/}.}
\conflictsofinterest{The authors declare no conflicts of interest.}
\useofartificialintelligence{OpenAI Codex (\texttt{gpt-5.5}, reasoning effort \texttt{xhigh}) assisted offline candidate-rule authoring, scripting, analysis, and drafting. The forecasting-time policy executes frozen JSON rules through a deterministic verifier. Appendix~\ref{app:template_induction_protocol} supplies the recovered task instruction, reference prompt, model configuration, and reproducibility materials.}

\par\medskip
\appendixtitles{yes}
\appendixstart
\appendix
\renewcommand{\theHtable}{appendix.\arabic{table}}
\section{Complete Seed-Level Activation Results}
\label{app:complete_results}

Table~\ref{tab:all_seed_results} reports all 543 dataset--horizon--configuration combinations and 1,471 unique training runs supporting this manuscript: 1,225 Informer, 198 PatchTST, and 48 iTransformer runs. The results are grouped by forecasting architecture and experimental scope. Each available seed is listed separately, and shared controls are included once. Table~\ref{tab:phase2_placement} presents representative comparisons from this full record.

The columns S1--S3 correspond to random seeds 2024--2026. A dash indicates that a run was not performed; $L/H$ denotes input length and prediction horizon. Plain configuration labels identify hidden-FFN activations. For Informer, enc-tanh and dec-tanh change only the encoder or decoder stack, respectively. The suffix +out adds tanh after the output projection; otherwise the output projection is linear. The tanh-sin perturbation amplitude is 0.01. Values are rounded to four decimal places; the accompanying CSV retains full precision, exact configuration identifiers, and source-file provenance.

\begingroup\footnotesize
\setlength{\tabcolsep}{3pt}
\renewcommand{\arraystretch}{1.05}
\begin{longtable}{@{}lllr rrrrrr@{}}
\caption{Complete activation results: individual-seed MSE and MAE (lower is better). Seeds S1, S2, and S3 are 2024, 2025, and 2026; a dash denotes an unobserved run.}\label{tab:all_seed_results}\\
\toprule Dataset & $L/H$ & Config. & & \multicolumn{2}{c}{S1} & \multicolumn{2}{c}{S2} & \multicolumn{2}{c}{S3}\\
& & & & MSE & MAE & MSE & MAE & MSE & MAE\\\midrule\endfirsthead
\caption[]{Complete activation results (continued)}\\\toprule Dataset & $L/H$ & Config. & & \multicolumn{2}{c}{S1} & \multicolumn{2}{c}{S2} & \multicolumn{2}{c}{S3}\\
& & & & MSE & MAE & MSE & MAE & MSE & MAE\\\midrule\endhead
\midrule\multicolumn{10}{r}{Continued on next page}\\\endfoot
\bottomrule\endlastfoot
\multicolumn{10}{l}{\textbf{Informer: complete grid}}\\*
ETTh1 & 48/24 & dec-tanh &  & 0.5394 & 0.5267 & 0.5428 & 0.5302 & 0.5671 & 0.5474\\
ETTh1 & 48/24 & enc-tanh &  & 0.5437 & 0.5337 & 0.5907 & 0.5563 & 0.5177 & 0.5159\\
ETTh1 & 48/24 & GELU &  & 0.5368 & 0.5253 & 0.6383 & 0.5824 & 0.5189 & 0.5164\\
ETTh1 & 48/24 & ReLU &  & 0.5449 & 0.5218 & 0.6683 & 0.5963 & 0.5355 & 0.5223\\
ETTh1 & 48/24 & softsign &  & 0.5057 & 0.5080 & 0.5568 & 0.5349 & 0.5347 & 0.5246\\
ETTh1 & 48/24 & Swish &  & 0.5141 & 0.5119 & 0.5758 & 0.5495 & 0.5783 & 0.5553\\
ETTh1 & 48/24 & tanh &  & 0.4975 & 0.4986 & 0.5593 & 0.5376 & 0.5423 & 0.5337\\
ETTh1 & 48/24 & tanh+out &  & 0.6201 & 0.5733 & 0.5976 & 0.5655 & 0.6058 & 0.5667\\
ETTh1 & 48/24 & tanh-sin &  & 0.5322 & 0.5120 & 0.5566 & 0.5358 & 0.5540 & 0.5424\\
ETTh1 & 96/48 & dec-tanh &  & 0.7512 & 0.6691 & 0.6907 & 0.6396 & 0.8819 & 0.7393\\
ETTh1 & 96/48 & enc-tanh &  & 0.6617 & 0.6173 & 0.6369 & 0.6144 & 0.8318 & 0.7066\\
ETTh1 & 96/48 & GELU &  & 0.7171 & 0.6479 & 0.7214 & 0.6502 & 0.7783 & 0.6910\\
ETTh1 & 96/48 & ReLU &  & 0.7564 & 0.6662 & 0.7614 & 0.6697 & 0.7551 & 0.6776\\
ETTh1 & 96/48 & softsign &  & 0.5617 & 0.5470 & 0.6831 & 0.6306 & 0.6929 & 0.6453\\
ETTh1 & 96/48 & Swish &  & 0.6901 & 0.6364 & 0.6403 & 0.6057 & 0.7544 & 0.6792\\
ETTh1 & 96/48 & tanh &  & 0.6049 & 0.5712 & 0.6945 & 0.6332 & 0.7032 & 0.6517\\
ETTh1 & 96/48 & tanh+out &  & 0.7660 & 0.6790 & 0.8028 & 0.7055 & 0.8166 & 0.7134\\
ETTh1 & 96/48 & tanh-sin &  & 0.5863 & 0.5563 & 0.6674 & 0.6214 & 0.7145 & 0.6506\\
ETTh1 & 168/168 & dec-tanh &  & 1.0272 & 0.8279 & 1.0718 & 0.8294 & 1.0685 & 0.8111\\
ETTh1 & 168/168 & enc-tanh &  & 0.9984 & 0.7941 & 1.1465 & 0.8324 & 0.9628 & 0.7827\\
ETTh1 & 168/168 & GELU &  & 1.0657 & 0.8295 & 1.0512 & 0.8369 & 1.0244 & 0.8104\\
ETTh1 & 168/168 & ReLU &  & 1.0804 & 0.8402 & 1.0658 & 0.8385 & 1.0438 & 0.8182\\
ETTh1 & 168/168 & softsign &  & 0.8545 & 0.7192 & 1.1019 & 0.8226 & 0.8417 & 0.7041\\
ETTh1 & 168/168 & Swish &  & 0.9905 & 0.7996 & 1.0508 & 0.8302 & 1.0242 & 0.8211\\
ETTh1 & 168/168 & tanh &  & 0.8308 & 0.7155 & 1.0624 & 0.8054 & 0.8415 & 0.7104\\
ETTh1 & 168/168 & tanh+out &  & 1.1038 & 0.8392 & 1.1206 & 0.8435 & 1.0927 & 0.8623\\
ETTh1 & 168/168 & tanh-sin &  & 0.8316 & 0.7146 & 1.0795 & 0.8159 & 0.8441 & 0.7118\\
ETTh1 & 168/336 & dec-tanh &  & 1.3968 & 0.9658 & 1.2093 & 0.8908 & 1.2612 & 0.9076\\
ETTh1 & 168/336 & enc-tanh &  & 1.3791 & 0.9435 & 1.4060 & 0.9576 & 1.3937 & 0.9639\\
ETTh1 & 168/336 & GELU &  & 1.3306 & 0.9224 & 1.3752 & 0.9595 & 1.3762 & 0.9619\\
ETTh1 & 168/336 & ReLU &  & 1.3760 & 0.9459 & 1.4501 & 1.0018 & 1.3071 & 0.9309\\
ETTh1 & 168/336 & softsign &  & 1.2651 & 0.9079 & 1.2087 & 0.8842 & 1.1831 & 0.8626\\
ETTh1 & 168/336 & Swish &  & 1.2908 & 0.9034 & 1.2540 & 0.9021 & 1.3428 & 0.9505\\
ETTh1 & 168/336 & tanh &  & 1.2958 & 0.9306 & 1.2460 & 0.8966 & 1.1469 & 0.8500\\
ETTh1 & 168/336 & tanh+out &  & 1.4416 & 1.0016 & 1.2123 & 0.8836 & 1.3677 & 0.9750\\
ETTh1 & 168/336 & tanh-sin &  & 1.2935 & 0.9328 & 1.2342 & 0.8901 & 1.1583 & 0.8550\\
ETTh1 & 336/720 & dec-tanh &  & 1.4746 & 1.0077 & 1.4305 & 0.9865 & 1.4143 & 0.9888\\
ETTh1 & 336/720 & enc-tanh &  & 1.5145 & 0.9920 & 1.4753 & 0.9846 & 1.4028 & 0.9623\\
ETTh1 & 336/720 & GELU &  & 1.3769 & 0.9390 & 1.4686 & 0.9883 & 1.3924 & 0.9710\\
ETTh1 & 336/720 & ReLU &  & 1.3430 & 0.9368 & 1.4102 & 0.9594 & 1.3385 & 0.9474\\
ETTh1 & 336/720 & softsign &  & 1.3235 & 0.9305 & 1.4339 & 0.9826 & 1.5187 & 1.0270\\
ETTh1 & 336/720 & Swish &  & 1.4479 & 0.9758 & 1.3869 & 0.9569 & 1.4255 & 0.9848\\
ETTh1 & 336/720 & tanh &  & 1.3994 & 0.9709 & 1.4758 & 1.0160 & 1.5320 & 1.0352\\
ETTh1 & 336/720 & tanh+out &  & 1.2958 & 0.9113 & 1.3555 & 0.9542 & 1.3322 & 0.9460\\
ETTh1 & 336/720 & tanh-sin &  & 1.4119 & 0.9767 & 1.4752 & 1.0171 & 1.5223 & 1.0315\\
ETTh2 & 48/24 & dec-tanh &  & 0.6079 & 0.6006 & 0.5819 & 0.5916 & 0.6109 & 0.5911\\
ETTh2 & 48/24 & enc-tanh &  & 0.6654 & 0.6313 & 0.6374 & 0.6186 & 0.6414 & 0.6109\\
ETTh2 & 48/24 & GELU &  & 0.6559 & 0.6325 & 0.8437 & 0.7241 & 0.6719 & 0.6240\\
ETTh2 & 48/24 & ReLU &  & 0.6520 & 0.6323 & 0.6289 & 0.6202 & 0.9142 & 0.7250\\
ETTh2 & 48/24 & softsign &  & 0.4918 & 0.5294 & 0.5470 & 0.5620 & 0.5657 & 0.5682\\
ETTh2 & 48/24 & Swish &  & 0.6033 & 0.5941 & 0.5804 & 0.5960 & 0.7459 & 0.6526\\
ETTh2 & 48/24 & tanh &  & 0.4887 & 0.5302 & 0.5353 & 0.5553 & 0.5641 & 0.5712\\
ETTh2 & 48/24 & tanh+out &  & 1.3871 & 0.9238 & 1.3545 & 0.8954 & 1.3684 & 0.9130\\
ETTh2 & 48/24 & tanh-sin &  & 0.4971 & 0.5320 & 0.5090 & 0.5431 & 0.5415 & 0.5537\\
ETTh2 & 96/48 & dec-tanh &  & 2.5997 & 1.3146 & 1.8963 & 1.1348 & 3.0971 & 1.4357\\
ETTh2 & 96/48 & enc-tanh &  & 2.7520 & 1.3625 & 1.9627 & 1.1719 & 2.8596 & 1.3925\\
ETTh2 & 96/48 & GELU &  & 2.7738 & 1.3662 & 2.1087 & 1.2204 & 2.9312 & 1.4053\\
ETTh2 & 96/48 & ReLU &  & 2.9465 & 1.3971 & 2.1314 & 1.1994 & 3.0506 & 1.4184\\
ETTh2 & 96/48 & softsign &  & 1.8608 & 1.1059 & 1.1626 & 0.8631 & 1.7960 & 1.0942\\
ETTh2 & 96/48 & Swish &  & 2.5686 & 1.3003 & 1.8128 & 1.1131 & 2.6597 & 1.3203\\
ETTh2 & 96/48 & tanh &  & 1.9433 & 1.1220 & 1.2907 & 0.9100 & 1.9251 & 1.1254\\
ETTh2 & 96/48 & tanh+out &  & 1.8048 & 1.1243 & 1.8425 & 1.1335 & 1.7101 & 1.0912\\
ETTh2 & 96/48 & tanh-sin &  & 1.8895 & 1.1115 & 1.2221 & 0.8872 & 1.8627 & 1.1099\\
ETTh2 & 336/168 & dec-tanh &  & 4.5220 & 1.7838 & 3.7536 & 1.5659 & 7.1384 & 2.2345\\
ETTh2 & 336/168 & enc-tanh &  & 3.1537 & 1.4590 & 2.6249 & 1.3655 & 3.6121 & 1.5361\\
ETTh2 & 336/168 & GELU &  & 4.8943 & 1.8629 & 4.1212 & 1.7539 & 6.4105 & 2.1213\\
ETTh2 & 336/168 & ReLU &  & 6.5460 & 2.2226 & 5.1756 & 1.9341 & 8.4577 & 2.4860\\
ETTh2 & 336/168 & softsign &  & 2.9250 & 1.3357 & 2.7532 & 1.2667 & 3.7564 & 1.4843\\
ETTh2 & 336/168 & Swish &  & 3.4420 & 1.5168 & 3.2483 & 1.4475 & 5.3205 & 1.8850\\
ETTh2 & 336/168 & tanh &  & 2.5559 & 1.2781 & 3.1508 & 1.3658 & 3.5263 & 1.4566\\
ETTh2 & 336/168 & tanh+out &  & 3.0216 & 1.3905 & 2.6107 & 1.3376 & 3.3568 & 1.4740\\
ETTh2 & 336/168 & tanh-sin &  & 2.4719 & 1.2573 & 2.9276 & 1.3265 & 3.4882 & 1.4389\\
ETTh2 & 336/336 & dec-tanh &  & 3.3865 & 1.4465 & 3.6320 & 1.5505 & 3.5009 & 1.5402\\
ETTh2 & 336/336 & enc-tanh &  & 2.7716 & 1.3116 & 3.3989 & 1.5233 & 2.8165 & 1.3762\\
ETTh2 & 336/336 & GELU &  & 4.1418 & 1.6369 & 3.7924 & 1.6220 & 3.8494 & 1.6440\\
ETTh2 & 336/336 & ReLU &  & 3.8754 & 1.5850 & 3.8505 & 1.6150 & 4.9302 & 1.8744\\
ETTh2 & 336/336 & softsign &  & 2.0891 & 1.1557 & 2.3708 & 1.2564 & 2.4814 & 1.2531\\
ETTh2 & 336/336 & Swish &  & 2.8878 & 1.3560 & 3.1786 & 1.4443 & 3.1543 & 1.4632\\
ETTh2 & 336/336 & tanh &  & 2.0336 & 1.1573 & 2.4406 & 1.2644 & 2.0604 & 1.1361\\
ETTh2 & 336/336 & tanh+out &  & 2.1627 & 1.1708 & 2.1969 & 1.1885 & 1.8021 & 1.0657\\
ETTh2 & 336/336 & tanh-sin &  & 2.0845 & 1.1664 & 2.3770 & 1.2504 & 2.0531 & 1.1368\\
ETTh2 & 720/720 & dec-tanh &  & 4.5175 & 1.8065 & 3.3406 & 1.5561 & 3.7857 & 1.5241\\
ETTh2 & 720/720 & enc-tanh &  & 4.4635 & 1.8443 & 3.3857 & 1.5246 & 3.3016 & 1.4953\\
ETTh2 & 720/720 & GELU &  & 4.1378 & 1.7754 & 3.6857 & 1.6063 & 4.4425 & 1.7404\\
ETTh2 & 720/720 & ReLU &  & 4.0503 & 1.7449 & 3.3610 & 1.5595 & 4.4921 & 1.7742\\
ETTh2 & 720/720 & softsign &  & 3.7966 & 1.6403 & 3.3566 & 1.5289 & 3.7741 & 1.5605\\
ETTh2 & 720/720 & Swish &  & 4.0527 & 1.7690 & 3.2707 & 1.5317 & 3.7440 & 1.5851\\
ETTh2 & 720/720 & tanh &  & 3.2243 & 1.4604 & 4.5994 & 1.7811 & 3.1291 & 1.3955\\
ETTh2 & 720/720 & tanh+out &  & 3.3753 & 1.4162 & 3.4333 & 1.5040 & 3.0860 & 1.3460\\
ETTh2 & 720/720 & tanh-sin &  & 3.2244 & 1.4569 & 4.5474 & 1.7848 & 3.0867 & 1.4030\\
ETTm1 & 96/48 & dec-tanh &  & 0.4712 & 0.4608 & 0.5031 & 0.4588 & 0.4804 & 0.4608\\
ETTm1 & 96/48 & enc-tanh &  & 0.4772 & 0.4723 & 0.4454 & 0.4640 & 0.4946 & 0.4699\\
ETTm1 & 96/48 & GELU &  & 0.4901 & 0.4614 & 0.5300 & 0.4803 & 0.5071 & 0.4873\\
ETTm1 & 96/48 & ReLU &  & 0.5345 & 0.4767 & 0.5171 & 0.4742 & 0.5113 & 0.4747\\
ETTm1 & 96/48 & softsign &  & 0.4304 & 0.4250 & 0.4799 & 0.4793 & 0.5247 & 0.4646\\
ETTm1 & 96/48 & Swish &  & 0.5163 & 0.4838 & 0.4949 & 0.4599 & 0.5000 & 0.4704\\
ETTm1 & 96/48 & tanh &  & 0.4424 & 0.4451 & 0.4381 & 0.4471 & 0.4795 & 0.4571\\
ETTm1 & 96/48 & tanh+out &  & 0.5404 & 0.5063 & 0.5711 & 0.5195 & 0.5526 & 0.5063\\
ETTm1 & 96/48 & tanh-sin &  & 0.4761 & 0.4572 & 0.4834 & 0.4564 & 0.4907 & 0.4467\\
ETTm1 & 384/96 & dec-tanh &  & 0.6638 & 0.6159 & 0.5889 & 0.5406 & 0.5716 & 0.5419\\
ETTm1 & 384/96 & enc-tanh &  & 0.6700 & 0.5963 & 0.6730 & 0.6061 & 0.6896 & 0.6112\\
ETTm1 & 384/96 & GELU &  & 0.6223 & 0.5653 & 0.6084 & 0.5617 & 0.6182 & 0.5686\\
ETTm1 & 384/96 & ReLU &  & 0.6257 & 0.5762 & 0.5880 & 0.5519 & 0.5923 & 0.5239\\
ETTm1 & 384/96 & softsign &  & 0.5854 & 0.5474 & 0.5181 & 0.5038 & 0.6124 & 0.5621\\
ETTm1 & 384/96 & Swish &  & 0.5888 & 0.5511 & 0.6065 & 0.5726 & 0.6160 & 0.5504\\
ETTm1 & 384/96 & tanh &  & 0.6333 & 0.5787 & 0.5442 & 0.5377 & 0.5732 & 0.5456\\
ETTm1 & 384/96 & tanh+out &  & 0.6787 & 0.6183 & 0.6352 & 0.5917 & 0.6192 & 0.5673\\
ETTm1 & 384/96 & tanh-sin &  & 0.6153 & 0.5847 & 0.5444 & 0.5308 & 0.5184 & 0.5032\\
ETTm1 & 672/24 & dec-tanh &  & 0.3296 & 0.3907 & 0.3364 & 0.3913 & 0.3199 & 0.3828\\
ETTm1 & 672/24 & enc-tanh &  & 0.3454 & 0.3943 & 0.3367 & 0.3844 & 0.3614 & 0.4048\\
ETTm1 & 672/24 & GELU &  & 0.3744 & 0.4058 & 0.3491 & 0.4075 & 0.3662 & 0.4166\\
ETTm1 & 672/24 & ReLU &  & 0.3858 & 0.4091 & 0.3365 & 0.3999 & 0.3540 & 0.4058\\
ETTm1 & 672/24 & softsign &  & 0.3389 & 0.3812 & 0.3048 & 0.3732 & 0.3547 & 0.3886\\
ETTm1 & 672/24 & Swish &  & 0.3356 & 0.3848 & 0.3258 & 0.3856 & 0.3345 & 0.3842\\
ETTm1 & 672/24 & tanh &  & 0.3059 & 0.3746 & 0.3055 & 0.3746 & 0.3320 & 0.3899\\
ETTm1 & 672/24 & tanh+out &  & 0.4219 & 0.4190 & 0.4681 & 0.4680 & 0.4120 & 0.4219\\
ETTm1 & 672/24 & tanh-sin &  & 0.3323 & 0.3871 & 0.3092 & 0.3688 & 0.3507 & 0.4000\\
ETTm1 & 672/288 & dec-tanh &  & 1.1258 & 0.8291 & 0.8833 & 0.7369 & 0.8637 & 0.7409\\
ETTm1 & 672/288 & enc-tanh &  & 0.8298 & 0.7175 & 0.9073 & 0.7312 & 0.9394 & 0.7790\\
ETTm1 & 672/288 & GELU &  & 1.1021 & 0.8160 & 0.8389 & 0.6918 & 0.7159 & 0.6367\\
ETTm1 & 672/288 & ReLU &  & 0.8713 & 0.7067 & 0.9120 & 0.7502 & 0.7967 & 0.6865\\
ETTm1 & 672/288 & softsign &  & 1.0387 & 0.8137 & 0.8486 & 0.7348 & 0.8151 & 0.7052\\
ETTm1 & 672/288 & Swish &  & 0.8272 & 0.7033 & 0.8627 & 0.7203 & 0.8711 & 0.7370\\
ETTm1 & 672/288 & tanh &  & 0.8344 & 0.7275 & 0.8193 & 0.7148 & 0.7521 & 0.6734\\
ETTm1 & 672/288 & tanh+out &  & 0.7274 & 0.6538 & 0.9745 & 0.7833 & 0.8117 & 0.6940\\
ETTm1 & 672/288 & tanh-sin &  & 0.8589 & 0.7408 & 0.7623 & 0.6875 & 0.7882 & 0.6909\\
ETTm1 & 672/672 & dec-tanh &  & 0.8832 & 0.7173 & 1.1765 & 0.8685 & 0.8221 & 0.6978\\
ETTm1 & 672/672 & enc-tanh &  & 0.9713 & 0.7771 & 1.0332 & 0.7970 & 0.9534 & 0.7483\\
ETTm1 & 672/672 & GELU &  & 0.9035 & 0.7131 & 1.0969 & 0.8053 & 1.0073 & 0.7821\\
ETTm1 & 672/672 & ReLU &  & 0.8724 & 0.7186 & 0.9920 & 0.7628 & 1.0328 & 0.7888\\
ETTm1 & 672/672 & softsign &  & 0.9473 & 0.7867 & 0.9011 & 0.7506 & 0.9177 & 0.7531\\
ETTm1 & 672/672 & Swish &  & 1.0396 & 0.7881 & 1.1161 & 0.8069 & 1.0724 & 0.8224\\
ETTm1 & 672/672 & tanh &  & 1.0374 & 0.8216 & 0.9928 & 0.8115 & 0.9619 & 0.7841\\
ETTm1 & 672/672 & tanh+out &  & 1.0558 & 0.8310 & 1.0309 & 0.8141 & 1.1004 & 0.8508\\
ETTm1 & 672/672 & tanh-sin &  & 1.0270 & 0.8214 & 0.9891 & 0.8096 & 1.0040 & 0.8058\\
ETTm2 & 96/48 & dec-tanh &  & 0.3265 & 0.4172 & 0.3041 & 0.3939 & 0.3739 & 0.4474\\
ETTm2 & 96/48 & enc-tanh &  & 0.3961 & 0.4651 & 0.3520 & 0.4299 & 0.4438 & 0.4856\\
ETTm2 & 96/48 & GELU &  & 0.4233 & 0.4771 & 0.3554 & 0.4230 & 0.4102 & 0.4554\\
ETTm2 & 96/48 & ReLU &  & 0.4475 & 0.4851 & 0.5033 & 0.5052 & 0.4842 & 0.5124\\
ETTm2 & 96/48 & softsign &  & 0.3013 & 0.3977 & 0.3014 & 0.3901 & 0.3249 & 0.4072\\
ETTm2 & 96/48 & Swish &  & 0.4641 & 0.4993 & 0.3230 & 0.4074 & 0.4353 & 0.4735\\
ETTm2 & 96/48 & tanh &  & 0.3288 & 0.4095 & 0.3374 & 0.4123 & 0.3394 & 0.4216\\
ETTm2 & 96/48 & tanh+out &  & 1.2554 & 0.7976 & 1.2636 & 0.8091 & 1.2561 & 0.8031\\
ETTm2 & 96/48 & tanh-sin &  & 0.3251 & 0.4099 & 0.2987 & 0.3904 & 0.3550 & 0.4326\\
ETTm2 & 384/96 & dec-tanh &  & 2.3636 & 1.2347 & 1.6795 & 1.0013 & 1.5382 & 1.0029\\
ETTm2 & 384/96 & enc-tanh &  & 2.1027 & 1.1080 & 1.5374 & 0.9845 & 1.4531 & 0.9684\\
ETTm2 & 384/96 & GELU &  & 1.9189 & 1.0889 & 1.5475 & 0.9913 & 1.3473 & 0.8969\\
ETTm2 & 384/96 & ReLU &  & 1.8000 & 1.0808 & 1.5990 & 1.0071 & 1.2565 & 0.8934\\
ETTm2 & 384/96 & softsign &  & 1.8002 & 1.0459 & 1.0956 & 0.8544 & 1.3131 & 0.8959\\
ETTm2 & 384/96 & Swish &  & 1.9855 & 1.1130 & 1.2022 & 0.8797 & 1.3385 & 0.9004\\
ETTm2 & 384/96 & tanh &  & 1.5113 & 0.9731 & 1.3601 & 0.9811 & 1.2436 & 0.8938\\
ETTm2 & 384/96 & tanh+out &  & 1.5749 & 1.0303 & 1.5572 & 1.0235 & 1.6611 & 1.0690\\
ETTm2 & 384/96 & tanh-sin &  & 1.5824 & 0.9778 & 1.2034 & 0.9073 & 1.2193 & 0.8832\\
ETTm2 & 672/24 & dec-tanh &  & 1.2415 & 0.8447 & 1.0338 & 0.7420 & 1.1196 & 0.7749\\
ETTm2 & 672/24 & enc-tanh &  & 1.1180 & 0.7761 & 1.3230 & 0.8320 & 1.4212 & 0.8905\\
ETTm2 & 672/24 & GELU &  & 1.0455 & 0.7718 & 1.3019 & 0.8809 & 1.3706 & 0.8749\\
ETTm2 & 672/24 & ReLU &  & 1.1878 & 0.8274 & 1.3272 & 0.8385 & 1.4465 & 0.9238\\
ETTm2 & 672/24 & softsign &  & 0.6347 & 0.5951 & 0.9730 & 0.7082 & 0.8383 & 0.6939\\
ETTm2 & 672/24 & Swish &  & 1.2896 & 0.8827 & 1.2972 & 0.8467 & 1.5460 & 0.9202\\
ETTm2 & 672/24 & tanh &  & 0.8934 & 0.7006 & 0.6432 & 0.5885 & 0.5174 & 0.5244\\
ETTm2 & 672/24 & tanh+out &  & 1.3195 & 0.8468 & 1.3184 & 0.8506 & 1.2546 & 0.7961\\
ETTm2 & 672/24 & tanh-sin &  & 0.8035 & 0.6774 & 0.9248 & 0.6981 & 0.5843 & 0.5637\\
ETTm2 & 672/288 & dec-tanh &  & 4.3824 & 1.8384 & 4.2616 & 1.7979 & 4.4833 & 1.7944\\
ETTm2 & 672/288 & enc-tanh &  & 3.5092 & 1.5121 & 3.2256 & 1.4933 & 4.1595 & 1.6500\\
ETTm2 & 672/288 & GELU &  & 5.7733 & 2.0660 & 4.2900 & 1.7790 & 5.9250 & 2.0651\\
ETTm2 & 672/288 & ReLU &  & 5.5451 & 2.0506 & 3.7592 & 1.6854 & 6.8824 & 2.2098\\
ETTm2 & 672/288 & softsign &  & 2.4419 & 1.2992 & 2.8449 & 1.4111 & 3.3410 & 1.5189\\
ETTm2 & 672/288 & Swish &  & 3.3248 & 1.5580 & 3.4238 & 1.5815 & 4.3682 & 1.7763\\
ETTm2 & 672/288 & tanh &  & 2.8916 & 1.3687 & 2.6157 & 1.3493 & 2.7990 & 1.3826\\
ETTm2 & 672/288 & tanh+out &  & 2.4563 & 1.3114 & 1.9555 & 1.1376 & 2.6519 & 1.3788\\
ETTm2 & 672/288 & tanh-sin &  & 2.4888 & 1.3039 & 2.4087 & 1.2968 & 2.6786 & 1.3512\\
ETTm2 & 672/672 & dec-tanh &  & 6.0446 & 2.1493 & 6.4759 & 2.2756 & 7.3546 & 2.3712\\
ETTm2 & 672/672 & enc-tanh &  & 5.8079 & 2.1045 & 4.7309 & 1.8487 & 6.7467 & 2.2181\\
ETTm2 & 672/672 & GELU &  & 6.5483 & 2.2452 & 8.7333 & 2.6607 & 8.1563 & 2.5493\\
ETTm2 & 672/672 & ReLU &  & 6.4909 & 2.2364 & 10.5802 & 2.9456 & 6.4135 & 2.2333\\
ETTm2 & 672/672 & softsign &  & 3.2860 & 1.5498 & 2.4866 & 1.3510 & 3.5921 & 1.6345\\
ETTm2 & 672/672 & Swish &  & 5.2692 & 2.0424 & 6.9000 & 2.3655 & 6.9140 & 2.3442\\
ETTm2 & 672/672 & tanh &  & 3.4973 & 1.5825 & 2.8441 & 1.4003 & 3.8157 & 1.6804\\
ETTm2 & 672/672 & tanh+out &  & 3.0320 & 1.4675 & 2.7200 & 1.3967 & 2.8007 & 1.4291\\
ETTm2 & 672/672 & tanh-sin &  & 3.3028 & 1.5416 & 2.5422 & 1.3394 & 3.3631 & 1.5672\\
Solar1 & 96/24 & dec-tanh &  & 0.1538 & 0.1642 & 0.1554 & 0.1684 & 0.1564 & 0.1648\\
Solar1 & 96/24 & enc-tanh &  & 0.1602 & 0.1695 & 0.1550 & 0.1641 & 0.1592 & 0.1661\\
Solar1 & 96/24 & GELU &  & 0.1594 & 0.1652 & 0.1476 & 0.1593 & 0.1525 & 0.1612\\
Solar1 & 96/24 & ReLU &  & 0.1590 & 0.1627 & 0.1543 & 0.1616 & 0.1574 & 0.1595\\
Solar1 & 96/24 & softsign &  & 0.1536 & 0.1711 & 0.1563 & 0.1721 & 0.1577 & 0.1664\\
Solar1 & 96/24 & Swish &  & 0.1504 & 0.1642 & 0.1508 & 0.1607 & 0.1502 & 0.1620\\
Solar1 & 96/24 & tanh &  & 0.1585 & 0.1687 & 0.1522 & 0.1729 & 0.1596 & 0.1697\\
Solar1 & 96/24 & tanh+out &  & 0.3771 & 0.2875 & 0.3705 & 0.2860 & 0.3733 & 0.2859\\
Solar1 & 96/24 & tanh-sin &  & 0.1591 & 0.1686 & 0.1514 & 0.1724 & 0.1524 & 0.1663\\
Solar1 & 96/96 & dec-tanh &  & 0.2329 & 0.2298 & 0.2294 & 0.2296 & 0.2414 & 0.2336\\
Solar1 & 96/96 & enc-tanh &  & 0.2283 & 0.2273 & 0.2186 & 0.2179 & 0.2254 & 0.2314\\
Solar1 & 96/96 & GELU &  & 0.2250 & 0.2225 & 0.2293 & 0.2264 & 0.2355 & 0.2296\\
Solar1 & 96/96 & ReLU &  & 0.2353 & 0.2236 & 0.2363 & 0.2251 & 0.2378 & 0.2268\\
Solar1 & 96/96 & softsign &  & 0.2246 & 0.2244 & 0.2242 & 0.2258 & 0.2228 & 0.2325\\
Solar1 & 96/96 & Swish &  & 0.2187 & 0.2216 & 0.2273 & 0.2329 & 0.2329 & 0.2346\\
Solar1 & 96/96 & tanh &  & 0.2171 & 0.2196 & 0.2331 & 0.2319 & 0.2241 & 0.2304\\
Solar1 & 96/96 & tanh+out &  & 0.4057 & 0.3185 & 0.4065 & 0.3187 & 0.4035 & 0.3167\\
Solar1 & 96/96 & tanh-sin &  & 0.2193 & 0.2216 & 0.2307 & 0.2298 & 0.2376 & 0.2408\\
Solar2 & 96/24 & dec-tanh &  & 0.0671 & 0.1314 & 0.0685 & 0.1353 & 0.0721 & 0.1362\\
Solar2 & 96/24 & enc-tanh &  & 0.0652 & 0.1254 & 0.0678 & 0.1311 & 0.0675 & 0.1293\\
Solar2 & 96/24 & GELU &  & 0.0725 & 0.1310 & 0.0685 & 0.1288 & 0.0710 & 0.1318\\
Solar2 & 96/24 & ReLU &  & 0.0662 & 0.1292 & 0.0717 & 0.1318 & 0.0721 & 0.1327\\
Solar2 & 96/24 & softsign &  & 0.0667 & 0.1320 & 0.0663 & 0.1316 & 0.0682 & 0.1363\\
Solar2 & 96/24 & Swish &  & 0.0682 & 0.1290 & 0.0653 & 0.1281 & 0.0679 & 0.1320\\
Solar2 & 96/24 & tanh &  & 0.0629 & 0.1285 & 0.0668 & 0.1328 & 0.0686 & 0.1345\\
Solar2 & 96/24 & tanh+out &  & 0.2048 & 0.2453 & 0.2071 & 0.2484 & 0.2050 & 0.2461\\
Solar2 & 96/24 & tanh-sin &  & 0.0656 & 0.1311 & 0.0658 & 0.1327 & 0.0671 & 0.1351\\
Solar2 & 96/96 & dec-tanh &  & 0.1210 & 0.1983 & 0.1183 & 0.1994 & 0.1190 & 0.1989\\
Solar2 & 96/96 & enc-tanh &  & 0.1052 & 0.1786 & 0.1173 & 0.1890 & 0.1055 & 0.1789\\
Solar2 & 96/96 & GELU &  & 0.1224 & 0.1923 & 0.1239 & 0.1932 & 0.1168 & 0.1911\\
Solar2 & 96/96 & ReLU &  & 0.1199 & 0.1930 & 0.1227 & 0.1925 & 0.1268 & 0.1997\\
Solar2 & 96/96 & softsign &  & 0.1149 & 0.1910 & 0.1134 & 0.1911 & 0.1085 & 0.1849\\
Solar2 & 96/96 & Swish &  & 0.1183 & 0.1906 & 0.1199 & 0.1928 & 0.1077 & 0.1839\\
Solar2 & 96/96 & tanh &  & 0.1095 & 0.1886 & 0.1115 & 0.1895 & 0.1029 & 0.1815\\
Solar2 & 96/96 & tanh+out &  & 0.2272 & 0.2748 & 0.2237 & 0.2717 & 0.2247 & 0.2755\\
Solar2 & 96/96 & tanh-sin &  & 0.1138 & 0.1906 & 0.1084 & 0.1923 & 0.1075 & 0.1855\\
Solar3 & 96/24 & dec-tanh &  & 0.2020 & 0.2715 & 0.2002 & 0.2680 & 0.2013 & 0.2679\\
Solar3 & 96/24 & enc-tanh &  & 0.2022 & 0.2603 & 0.2006 & 0.2615 & 0.2008 & 0.2626\\
Solar3 & 96/24 & GELU &  & 0.1962 & 0.2593 & 0.1892 & 0.2548 & 0.2072 & 0.2632\\
Solar3 & 96/24 & ReLU &  & 0.1998 & 0.2582 & 0.1942 & 0.2555 & 0.2042 & 0.2610\\
Solar3 & 96/24 & softsign &  & 0.2051 & 0.2657 & 0.2028 & 0.2680 & 0.2041 & 0.2674\\
Solar3 & 96/24 & Swish &  & 0.1956 & 0.2597 & 0.2036 & 0.2634 & 0.1984 & 0.2599\\
Solar3 & 96/24 & tanh &  & 0.2061 & 0.2817 & 0.1920 & 0.2656 & 0.1992 & 0.2703\\
Solar3 & 96/24 & tanh+out &  & 0.3864 & 0.3966 & 0.3869 & 0.3964 & 0.3907 & 0.4001\\
Solar3 & 96/24 & tanh-sin &  & 0.2012 & 0.2774 & 0.2002 & 0.2704 & 0.1974 & 0.2692\\
Solar3 & 96/96 & dec-tanh &  & 0.3671 & 0.3815 & 0.3479 & 0.3740 & 0.3640 & 0.3825\\
Solar3 & 96/96 & enc-tanh &  & 0.3819 & 0.3888 & 0.3512 & 0.3713 & 0.3515 & 0.3719\\
Solar3 & 96/96 & GELU &  & 0.3415 & 0.3560 & 0.3556 & 0.3667 & 0.3499 & 0.3622\\
Solar3 & 96/96 & ReLU &  & 0.3627 & 0.3645 & 0.3438 & 0.3534 & 0.3666 & 0.3729\\
Solar3 & 96/96 & softsign &  & 0.3651 & 0.3897 & 0.3533 & 0.3769 & 0.3541 & 0.3871\\
Solar3 & 96/96 & Swish &  & 0.3535 & 0.3659 & 0.3716 & 0.3723 & 0.3543 & 0.3690\\
Solar3 & 96/96 & tanh &  & 0.3952 & 0.3989 & 0.3336 & 0.3781 & 0.3866 & 0.4029\\
Solar3 & 96/96 & tanh+out &  & 0.4482 & 0.4434 & 0.4700 & 0.4632 & 0.4535 & 0.4484\\
Solar3 & 96/96 & tanh-sin &  & 0.3559 & 0.3816 & 0.3498 & 0.3882 & 0.3692 & 0.3958\\
Solar4 & 96/24 & dec-tanh &  & 0.2218 & 0.2706 & 0.2226 & 0.2717 & 0.2252 & 0.2729\\
Solar4 & 96/24 & enc-tanh &  & 0.2213 & 0.2672 & 0.2225 & 0.2691 & 0.2237 & 0.2685\\
Solar4 & 96/24 & GELU &  & 0.2253 & 0.2675 & 0.2221 & 0.2659 & 0.2251 & 0.2687\\
Solar4 & 96/24 & ReLU &  & 0.2230 & 0.2648 & 0.2253 & 0.2667 & 0.2241 & 0.2670\\
Solar4 & 96/24 & softsign &  & 0.2175 & 0.2670 & 0.2206 & 0.2706 & 0.2221 & 0.2712\\
Solar4 & 96/24 & Swish &  & 0.2190 & 0.2650 & 0.2225 & 0.2689 & 0.2188 & 0.2659\\
Solar4 & 96/24 & tanh &  & 0.2219 & 0.2712 & 0.2209 & 0.2711 & 0.2260 & 0.2721\\
Solar4 & 96/24 & tanh+out &  & 0.3466 & 0.3421 & 0.3464 & 0.3430 & 0.3420 & 0.3410\\
Solar4 & 96/24 & tanh-sin &  & 0.2169 & 0.2675 & 0.2211 & 0.2714 & 0.2214 & 0.2704\\
Solar4 & 96/96 & dec-tanh &  & 0.3003 & 0.3186 & 0.2952 & 0.3190 & 0.2978 & 0.3162\\
Solar4 & 96/96 & enc-tanh &  & 0.2899 & 0.3163 & 0.2861 & 0.3166 & 0.2953 & 0.3196\\
Solar4 & 96/96 & GELU &  & 0.2978 & 0.3139 & 0.2990 & 0.3146 & 0.2920 & 0.3132\\
Solar4 & 96/96 & ReLU &  & 0.3032 & 0.3156 & 0.3003 & 0.3132 & 0.3009 & 0.3166\\
Solar4 & 96/96 & softsign &  & 0.2927 & 0.3171 & 0.2977 & 0.3215 & 0.2913 & 0.3137\\
Solar4 & 96/96 & Swish &  & 0.3037 & 0.3210 & 0.2949 & 0.3171 & 0.2962 & 0.3179\\
Solar4 & 96/96 & tanh &  & 0.2938 & 0.3203 & 0.2947 & 0.3187 & 0.2897 & 0.3164\\
Solar4 & 96/96 & tanh+out &  & 0.3912 & 0.3747 & 0.3930 & 0.3748 & 0.3920 & 0.3757\\
Solar4 & 96/96 & tanh-sin &  & 0.3052 & 0.3257 & 0.2914 & 0.3158 & 0.2886 & 0.3170\\
Solar5 & 96/24 & dec-tanh &  & 0.1071 & 0.1583 & 0.1124 & 0.1609 & 0.1145 & 0.1636\\
Solar5 & 96/24 & enc-tanh &  & 0.1119 & 0.1580 & 0.1097 & 0.1564 & 0.1138 & 0.1608\\
Solar5 & 96/24 & GELU &  & 0.1119 & 0.1572 & 0.1109 & 0.1566 & 0.1176 & 0.1619\\
Solar5 & 96/24 & ReLU &  & 0.1142 & 0.1615 & 0.1177 & 0.1613 & 0.1167 & 0.1627\\
Solar5 & 96/24 & softsign &  & 0.1144 & 0.1639 & 0.1098 & 0.1607 & 0.1062 & 0.1581\\
Solar5 & 96/24 & Swish &  & 0.1067 & 0.1555 & 0.1087 & 0.1551 & 0.1107 & 0.1590\\
Solar5 & 96/24 & tanh &  & 0.1073 & 0.1593 & 0.1135 & 0.1608 & 0.1083 & 0.1621\\
Solar5 & 96/24 & tanh+out &  & 0.2290 & 0.2466 & 0.2277 & 0.2439 & 0.2259 & 0.2467\\
Solar5 & 96/24 & tanh-sin &  & 0.1131 & 0.1645 & 0.1113 & 0.1614 & 0.1092 & 0.1623\\
Solar5 & 96/96 & dec-tanh &  & 0.1912 & 0.2303 & 0.1939 & 0.2334 & 0.1967 & 0.2399\\
Solar5 & 96/96 & enc-tanh &  & 0.1776 & 0.2212 & 0.1827 & 0.2266 & 0.1809 & 0.2312\\
Solar5 & 96/96 & GELU &  & 0.1935 & 0.2297 & 0.2010 & 0.2362 & 0.1913 & 0.2330\\
Solar5 & 96/96 & ReLU &  & 0.1910 & 0.2275 & 0.2030 & 0.2359 & 0.2008 & 0.2364\\
Solar5 & 96/96 & softsign &  & 0.1830 & 0.2261 & 0.1783 & 0.2273 & 0.1865 & 0.2277\\
Solar5 & 96/96 & Swish &  & 0.1912 & 0.2291 & 0.1949 & 0.2330 & 0.1953 & 0.2356\\
Solar5 & 96/96 & tanh &  & 0.1893 & 0.2264 & 0.1825 & 0.2305 & 0.1869 & 0.2276\\
Solar5 & 96/96 & tanh+out &  & 0.2652 & 0.2848 & 0.2653 & 0.2856 & 0.2634 & 0.2862\\
Solar5 & 96/96 & tanh-sin &  & 0.1794 & 0.2212 & 0.1790 & 0.2272 & 0.1845 & 0.2302\\
Solar6 & 96/24 & dec-tanh &  & 0.2652 & 0.1787 & 0.2688 & 0.1877 & 0.2675 & 0.1834\\
Solar6 & 96/24 & enc-tanh &  & 0.2676 & 0.1779 & 0.2691 & 0.1834 & 0.2596 & 0.1769\\
Solar6 & 96/24 & GELU &  & 0.2678 & 0.1761 & 0.2711 & 0.1834 & 0.2645 & 0.1786\\
Solar6 & 96/24 & ReLU &  & 0.2692 & 0.1758 & 0.2686 & 0.1781 & 0.2663 & 0.1757\\
Solar6 & 96/24 & softsign &  & 0.2699 & 0.1819 & 0.2687 & 0.1849 & 0.2703 & 0.1865\\
Solar6 & 96/24 & Swish &  & 0.2621 & 0.1753 & 0.2673 & 0.1831 & 0.2679 & 0.1836\\
Solar6 & 96/24 & tanh &  & 0.2726 & 0.1843 & 0.2719 & 0.1893 & 0.2656 & 0.1812\\
Solar6 & 96/24 & tanh+out &  & 0.3272 & 0.2437 & 0.3277 & 0.2468 & 0.3265 & 0.2438\\
Solar6 & 96/24 & tanh-sin &  & 0.2666 & 0.1809 & 0.2642 & 0.1820 & 0.2635 & 0.1792\\
Solar6 & 96/96 & dec-tanh &  & 0.3722 & 0.2684 & 0.3699 & 0.2629 & 0.3582 & 0.2602\\
Solar6 & 96/96 & enc-tanh &  & 0.3632 & 0.2614 & 0.3546 & 0.2572 & 0.3602 & 0.2618\\
Solar6 & 96/96 & GELU &  & 0.3636 & 0.2593 & 0.3680 & 0.2608 & 0.3660 & 0.2653\\
Solar6 & 96/96 & ReLU &  & 0.3674 & 0.2621 & 0.3596 & 0.2559 & 0.3632 & 0.2602\\
Solar6 & 96/96 & softsign &  & 0.3647 & 0.2602 & 0.3539 & 0.2555 & 0.3684 & 0.2628\\
Solar6 & 96/96 & Swish &  & 0.3694 & 0.2653 & 0.3671 & 0.2630 & 0.3496 & 0.2545\\
Solar6 & 96/96 & tanh &  & 0.3607 & 0.2594 & 0.3585 & 0.2575 & 0.3632 & 0.2633\\
Solar6 & 96/96 & tanh+out &  & 0.3846 & 0.2981 & 0.3792 & 0.2930 & 0.3814 & 0.2966\\
Solar6 & 96/96 & tanh-sin &  & 0.3621 & 0.2594 & 0.3468 & 0.2530 & 0.3639 & 0.2644\\
Solar7 & 96/24 & dec-tanh &  & 3.9492 & 1.3139 & 5.1303 & 1.5083 & 3.9411 & 1.2962\\
Solar7 & 96/24 & enc-tanh &  & 4.6774 & 1.2824 & 3.4664 & 1.2394 & 2.5613 & 1.1228\\
Solar7 & 96/24 & GELU &  & 2.9334 & 1.1211 & 3.0891 & 1.1205 & 5.5613 & 1.6230\\
Solar7 & 96/24 & ReLU &  & 3.4565 & 1.3380 & 2.1692 & 1.0539 & 4.2165 & 1.3536\\
Solar7 & 96/24 & softsign &  & 4.7487 & 1.4068 & 5.0490 & 1.4138 & 4.8680 & 1.4345\\
Solar7 & 96/24 & Swish &  & 4.6760 & 1.3590 & 4.6324 & 1.4382 & 3.6340 & 1.2393\\
Solar7 & 96/24 & tanh &  & 3.8714 & 1.2636 & 3.8074 & 1.2876 & 3.8066 & 1.2622\\
Solar7 & 96/24 & tanh+out &  & 10.6836 & 1.9277 & 10.8627 & 2.0142 & 10.8920 & 2.0398\\
Solar7 & 96/24 & tanh-sin &  & 4.9367 & 1.4160 & 4.8726 & 1.3885 & 4.1529 & 1.3175\\
Solar7 & 96/96 & dec-tanh &  & 6.2790 & 1.5100 & 7.8850 & 1.7634 & 6.8148 & 1.6702\\
Solar7 & 96/96 & enc-tanh &  & 6.5281 & 1.4380 & 4.2878 & 1.3233 & 4.0930 & 1.2478\\
Solar7 & 96/96 & GELU &  & 8.8628 & 1.7520 & 7.1642 & 1.7418 & 7.8817 & 1.7290\\
Solar7 & 96/96 & ReLU &  & 9.7201 & 1.7790 & 9.5950 & 1.7695 & 7.2110 & 1.6322\\
Solar7 & 96/96 & softsign &  & 5.1484 & 1.2994 & 5.3515 & 1.4338 & 6.7927 & 1.5690\\
Solar7 & 96/96 & Swish &  & 8.2039 & 1.6168 & 8.0854 & 1.6786 & 5.5964 & 1.4597\\
Solar7 & 96/96 & tanh &  & 6.9133 & 1.4903 & 6.1886 & 1.5632 & 5.9194 & 1.4877\\
Solar7 & 96/96 & tanh+out &  & 10.5271 & 1.7037 & 10.4746 & 1.7526 & 10.5101 & 1.7815\\
Solar7 & 96/96 & tanh-sin &  & 6.9795 & 1.4826 & 6.2304 & 1.6089 & 5.0917 & 1.4171\\
Solar8 & 96/24 & dec-tanh &  & 0.2159 & 0.2696 & 0.2163 & 0.2675 & 0.2195 & 0.2705\\
Solar8 & 96/24 & enc-tanh &  & 0.2148 & 0.2666 & 0.2171 & 0.2677 & 0.2166 & 0.2666\\
Solar8 & 96/24 & GELU &  & 0.2161 & 0.2648 & 0.2191 & 0.2662 & 0.2177 & 0.2656\\
Solar8 & 96/24 & ReLU &  & 0.2199 & 0.2656 & 0.2197 & 0.2664 & 0.2202 & 0.2665\\
Solar8 & 96/24 & softsign &  & 0.2172 & 0.2695 & 0.2185 & 0.2707 & 0.2161 & 0.2670\\
Solar8 & 96/24 & Swish &  & 0.2182 & 0.2678 & 0.2162 & 0.2654 & 0.2171 & 0.2661\\
Solar8 & 96/24 & tanh &  & 0.2140 & 0.2673 & 0.2140 & 0.2675 & 0.2212 & 0.2723\\
Solar8 & 96/24 & tanh+out &  & 0.3379 & 0.3437 & 0.3381 & 0.3437 & 0.3383 & 0.3432\\
Solar8 & 96/24 & tanh-sin &  & 0.2142 & 0.2665 & 0.2189 & 0.2734 & 0.2149 & 0.2667\\
Solar8 & 96/96 & dec-tanh &  & 0.3175 & 0.3418 & 0.3146 & 0.3407 & 0.3055 & 0.3324\\
Solar8 & 96/96 & enc-tanh &  & 0.3155 & 0.3432 & 0.3183 & 0.3424 & 0.3127 & 0.3361\\
Solar8 & 96/96 & GELU &  & 0.3151 & 0.3377 & 0.3147 & 0.3374 & 0.3115 & 0.3349\\
Solar8 & 96/96 & ReLU &  & 0.3190 & 0.3376 & 0.3181 & 0.3383 & 0.3061 & 0.3312\\
Solar8 & 96/96 & softsign &  & 0.3106 & 0.3363 & 0.3002 & 0.3316 & 0.3060 & 0.3355\\
Solar8 & 96/96 & Swish &  & 0.3140 & 0.3377 & 0.3145 & 0.3371 & 0.3105 & 0.3360\\
Solar8 & 96/96 & tanh &  & 0.3137 & 0.3412 & 0.3155 & 0.3401 & 0.3048 & 0.3337\\
Solar8 & 96/96 & tanh+out &  & 0.3873 & 0.3811 & 0.3931 & 0.3893 & 0.3907 & 0.3850\\
Solar8 & 96/96 & tanh-sin &  & 0.3089 & 0.3395 & 0.3084 & 0.3341 & 0.3095 & 0.3381\\
\multicolumn{10}{l}{\textbf{Informer: additional datasets}}\\*
Exchange & 96/96 & dec-tanh &  & 0.9305 & 0.7732 & 0.9218 & 0.7560 & 0.9774 & 0.7854\\
Exchange & 96/96 & enc-tanh &  & 1.0744 & 0.8360 & 1.0338 & 0.8141 & 1.1410 & 0.8597\\
Exchange & 96/96 & GELU &  & 0.9701 & 0.7881 & 0.9497 & 0.7753 & 1.0113 & 0.7997\\
Exchange & 96/96 & softsign &  & 1.0053 & 0.8089 & 1.0410 & 0.8059 & 1.0725 & 0.8296\\
Exchange & 96/96 & tanh &  & 1.0143 & 0.8111 & 1.0031 & 0.7910 & 1.0643 & 0.8272\\
Exchange & 96/96 & tanh+out &  & 1.2902 & 0.9091 & 1.2666 & 0.8946 & 1.2580 & 0.8857\\
Exchange & 96/96 & tanh-sin &  & 1.0121 & 0.8099 & 1.0042 & 0.7915 & 1.0645 & 0.8275\\
Exchange & 96/192 & dec-tanh &  & 1.1411 & 0.8620 & 1.0841 & 0.8307 & 1.1718 & 0.8602\\
Exchange & 96/192 & enc-tanh &  & 1.5742 & 1.0260 & 1.4836 & 0.9885 & 1.4949 & 0.9804\\
Exchange & 96/192 & GELU &  & 1.1813 & 0.8699 & 1.1027 & 0.8330 & 1.1629 & 0.8608\\
Exchange & 96/192 & softsign &  & 1.6034 & 1.0234 & 1.4022 & 0.9523 & 1.4495 & 0.9602\\
Exchange & 96/192 & tanh &  & 1.6125 & 1.0341 & 1.4763 & 0.9799 & 1.4773 & 0.9733\\
Exchange & 96/192 & tanh+out &  & 1.5865 & 1.0220 & 1.5354 & 1.0001 & 1.5250 & 0.9858\\
Exchange & 96/192 & tanh-sin &  & 1.6146 & 1.0348 & 1.4752 & 0.9797 & 1.4760 & 0.9731\\
Exchange & 96/336 & dec-tanh &  & 1.6929 & 1.0286 & 1.4957 & 0.9529 & 1.7073 & 1.0388\\
Exchange & 96/336 & enc-tanh &  & 1.8732 & 1.1049 & 1.8034 & 1.0838 & 1.9560 & 1.1313\\
Exchange & 96/336 & GELU &  & 1.6637 & 1.0199 & 1.5585 & 0.9839 & 1.7111 & 1.0436\\
Exchange & 96/336 & softsign &  & 1.7164 & 1.0697 & 1.7022 & 1.0522 & 1.8138 & 1.0872\\
Exchange & 96/336 & tanh &  & 1.7793 & 1.0927 & 1.7536 & 1.0647 & 1.8481 & 1.1020\\
Exchange & 96/336 & tanh+out &  & 1.7208 & 1.0593 & 1.7707 & 1.0722 & 1.8430 & 1.0939\\
Exchange & 96/336 & tanh-sin &  & 1.7785 & 1.0930 & 1.7491 & 1.0634 & 1.8455 & 1.1014\\
ILI & 36/24 & dec-tanh &  & 5.8530 & 1.6455 & 5.4854 & 1.5964 & 6.0159 & 1.6956\\
ILI & 36/24 & enc-tanh &  & 5.9863 & 1.6866 & 5.5242 & 1.6048 & 5.8697 & 1.6593\\
ILI & 36/24 & GELU &  & 5.8972 & 1.6576 & 5.5255 & 1.6018 & 6.0731 & 1.7049\\
ILI & 36/24 & softsign &  & 5.9170 & 1.6683 & 5.4781 & 1.5959 & 5.8834 & 1.6619\\
ILI & 36/24 & tanh &  & 5.9111 & 1.6685 & 5.4251 & 1.5843 & 5.8281 & 1.6489\\
ILI & 36/24 & tanh+out &  & 6.0060 & 1.7141 & 5.8315 & 1.6809 & 5.8584 & 1.6801\\
ILI & 36/24 & tanh-sin &  & 5.8858 & 1.6615 & 5.4242 & 1.5847 & 5.8256 & 1.6475\\
ILI & 36/36 & dec-tanh &  & 5.6881 & 1.6689 & 5.8394 & 1.6701 & 5.5897 & 1.6241\\
ILI & 36/36 & enc-tanh &  & 5.4236 & 1.6120 & 5.5913 & 1.6218 & 5.4359 & 1.5890\\
ILI & 36/36 & GELU &  & 5.6819 & 1.6655 & 5.8244 & 1.6662 & 5.6323 & 1.6336\\
ILI & 36/36 & softsign &  & 5.5256 & 1.6364 & 5.7289 & 1.6433 & 5.3917 & 1.5751\\
ILI & 36/36 & tanh &  & 5.5155 & 1.6336 & 5.6108 & 1.6218 & 5.4108 & 1.5820\\
ILI & 36/36 & tanh+out &  & 5.6826 & 1.6695 & 5.9127 & 1.7025 & 5.6655 & 1.6558\\
ILI & 36/36 & tanh-sin &  & 5.4970 & 1.6299 & 5.5974 & 1.6197 & 5.4133 & 1.5823\\
ILI & 36/48 & dec-tanh &  & 5.1635 & 1.5532 & 5.2954 & 1.5755 & 5.0428 & 1.5173\\
ILI & 36/48 & enc-tanh &  & 4.9708 & 1.5217 & 5.1024 & 1.5446 & 5.0160 & 1.5170\\
ILI & 36/48 & GELU &  & 5.1766 & 1.5559 & 5.2692 & 1.5746 & 5.1470 & 1.5432\\
ILI & 36/48 & softsign &  & 4.9304 & 1.5082 & 5.2340 & 1.5714 & 5.0112 & 1.5089\\
ILI & 36/48 & tanh &  & 4.9051 & 1.5034 & 5.1749 & 1.5579 & 4.9817 & 1.5020\\
ILI & 36/48 & tanh+out &  & 5.4433 & 1.6048 & 5.4937 & 1.6133 & 5.4802 & 1.6043\\
ILI & 36/48 & tanh-sin &  & 4.8987 & 1.5021 & 5.1808 & 1.5591 & 5.0043 & 1.5056\\
Weather & 96/96 & dec-tanh &  & 0.4430 & 0.4765 & 0.3826 & 0.4214 & 0.4122 & 0.4508\\
Weather & 96/96 & enc-tanh &  & 0.4500 & 0.4779 & 0.3788 & 0.4326 & 0.3772 & 0.4331\\
Weather & 96/96 & GELU &  & 0.4191 & 0.4517 & 0.3300 & 0.3876 & 0.3514 & 0.4040\\
Weather & 96/96 & softsign &  & 0.4045 & 0.4524 & 0.3614 & 0.4222 & 0.3596 & 0.4244\\
Weather & 96/96 & tanh &  & 0.3478 & 0.4105 & 0.3521 & 0.4125 & 0.3548 & 0.4158\\
Weather & 96/96 & tanh+out &  & 0.4032 & 0.4411 & 0.3355 & 0.3976 & 0.3595 & 0.4132\\
Weather & 96/96 & tanh-sin &  & 0.3369 & 0.4008 & 0.3556 & 0.4161 & 0.3545 & 0.4168\\
Weather & 96/192 & dec-tanh &  & 0.3815 & 0.4113 & 0.3950 & 0.4199 & 0.4283 & 0.4434\\
Weather & 96/192 & enc-tanh &  & 0.4866 & 0.4857 & 0.4853 & 0.4820 & 0.5025 & 0.4926\\
Weather & 96/192 & GELU &  & 0.3718 & 0.4054 & 0.3897 & 0.4138 & 0.4465 & 0.4522\\
Weather & 96/192 & softsign &  & 0.5147 & 0.4965 & 0.5147 & 0.5005 & 0.5139 & 0.5075\\
Weather & 96/192 & tanh &  & 0.5122 & 0.4964 & 0.5329 & 0.5117 & 0.5256 & 0.5112\\
Weather & 96/192 & tanh+out &  & 0.5040 & 0.4865 & 0.5282 & 0.5004 & 0.4901 & 0.4840\\
Weather & 96/192 & tanh-sin &  & 0.4941 & 0.4829 & 0.5054 & 0.4935 & 0.5137 & 0.5014\\
Weather & 96/336 & dec-tanh &  & 0.4869 & 0.4761 & 0.4695 & 0.4595 & 0.4267 & 0.4309\\
Weather & 96/336 & enc-tanh &  & 0.6188 & 0.5532 & 0.6344 & 0.5704 & 0.5741 & 0.5308\\
Weather & 96/336 & GELU &  & 0.4643 & 0.4696 & 0.4914 & 0.4833 & 0.4466 & 0.4523\\
Weather & 96/336 & softsign &  & 0.7291 & 0.6076 & 0.6838 & 0.5881 & 0.6832 & 0.5892\\
Weather & 96/336 & tanh &  & 0.6539 & 0.5640 & 0.6439 & 0.5667 & 0.7128 & 0.6007\\
Weather & 96/336 & tanh+out &  & 0.6889 & 0.5874 & 0.6226 & 0.5454 & 0.6968 & 0.5886\\
Weather & 96/336 & tanh-sin &  & 0.6599 & 0.5702 & 0.6550 & 0.5742 & 0.6792 & 0.5793\\
\multicolumn{10}{l}{\textbf{Informer: held-out and confirmation}}\\*
Electricity & 96/96 & dec-tanh &  & 0.2773 & 0.3747 & 0.2668 & 0.3652 & 0.2783 & 0.3733\\
Electricity & 96/96 & enc-tanh &  & 0.2948 & 0.3856 & -- & -- & -- & --\\
Electricity & 96/96 & GELU &  & 0.2817 & 0.3762 & 0.2730 & 0.3685 & 0.2782 & 0.3718\\
Electricity & 96/96 & softsign &  & 0.2900 & 0.3854 & -- & -- & -- & --\\
Electricity & 96/96 & tanh &  & 0.2928 & 0.3869 & -- & -- & -- & --\\
Electricity & 96/96 & tanh+out &  & 0.3290 & 0.4089 & -- & -- & -- & --\\
Electricity & 96/96 & tanh-sin &  & 0.2914 & 0.3848 & -- & -- & -- & --\\
Electricity & 96/192 & dec-tanh &  & 0.2839 & 0.3788 & -- & -- & -- & --\\
Electricity & 96/192 & enc-tanh &  & 0.2879 & 0.3839 & -- & -- & -- & --\\
Electricity & 96/192 & GELU &  & 0.2812 & 0.3758 & -- & -- & -- & --\\
Electricity & 96/192 & softsign &  & 0.2798 & 0.3781 & -- & -- & -- & --\\
Electricity & 96/192 & tanh &  & 0.2791 & 0.3778 & -- & -- & -- & --\\
Electricity & 96/192 & tanh+out &  & 0.3308 & 0.4138 & -- & -- & -- & --\\
Electricity & 96/192 & tanh-sin &  & 0.2792 & 0.3780 & -- & -- & -- & --\\
Electricity & 96/336 & dec-tanh &  & 0.3076 & 0.4016 & 0.3160 & 0.4092 & 0.2797 & 0.3784\\
Electricity & 96/336 & enc-tanh &  & 0.3618 & 0.4319 & -- & -- & -- & --\\
Electricity & 96/336 & GELU &  & 0.3080 & 0.4007 & 0.3092 & 0.4036 & 0.2917 & 0.3863\\
Electricity & 96/336 & softsign &  & 0.3210 & 0.4165 & -- & -- & -- & --\\
Electricity & 96/336 & tanh &  & 0.3340 & 0.4233 & -- & -- & -- & --\\
Electricity & 96/336 & tanh+out &  & 0.3520 & 0.4337 & -- & -- & -- & --\\
Electricity & 96/336 & tanh-sin &  & 0.3332 & 0.4232 & -- & -- & -- & --\\
Traffic & 96/96 & dec-tanh &  & 0.7226 & 0.3951 & -- & -- & -- & --\\
Traffic & 96/96 & enc-tanh &  & 0.7143 & 0.3981 & -- & -- & -- & --\\
Traffic & 96/96 & GELU &  & 0.7011 & 0.3903 & 0.7286 & 0.3924 & 0.6781 & 0.3757\\
Traffic & 96/96 & softsign &  & 0.7255 & 0.4046 & -- & -- & -- & --\\
Traffic & 96/96 & tanh &  & 0.7154 & 0.3977 & -- & -- & -- & --\\
Traffic & 96/96 & tanh+out &  & 0.8073 & 0.4239 & -- & -- & -- & --\\
Traffic & 96/96 & tanh-sin &  & 0.7068 & 0.3945 & 0.7068 & 0.3916 & 0.6756 & 0.3746\\
Traffic & 96/192 & dec-tanh &  & 0.7353 & 0.4005 & -- & -- & -- & --\\
Traffic & 96/192 & enc-tanh &  & 0.7476 & 0.4093 & -- & -- & -- & --\\
Traffic & 96/192 & GELU &  & 0.7253 & 0.3946 & 0.7674 & 0.4143 & 0.7577 & 0.4048\\
Traffic & 96/192 & softsign &  & 0.7429 & 0.4049 & -- & -- & -- & --\\
Traffic & 96/192 & tanh &  & 0.7545 & 0.4077 & -- & -- & -- & --\\
Traffic & 96/192 & tanh+out &  & 0.8304 & 0.4285 & -- & -- & -- & --\\
Traffic & 96/192 & tanh-sin &  & 0.7494 & 0.4047 & 0.7798 & 0.4230 & 0.7430 & 0.3982\\
Traffic & 96/336 & dec-tanh &  & 0.7564 & 0.4162 & -- & -- & -- & --\\
Traffic & 96/336 & enc-tanh &  & 0.7456 & 0.4127 & -- & -- & -- & --\\
Traffic & 96/336 & GELU &  & 0.7667 & 0.4218 & 0.8186 & 0.4426 & 0.7525 & 0.4112\\
Traffic & 96/336 & softsign &  & 0.7360 & 0.4057 & 0.8330 & 0.4551 & 0.7662 & 0.4176\\
Traffic & 96/336 & tanh &  & 0.7384 & 0.4063 & -- & -- & -- & --\\
Traffic & 96/336 & tanh+out &  & 0.8497 & 0.4485 & -- & -- & -- & --\\
Traffic & 96/336 & tanh-sin &  & 0.7363 & 0.4067 & 0.8089 & 0.4390 & 0.7749 & 0.4170\\
\multicolumn{10}{l}{\textbf{PatchTST: regime and output placement}}\\*
ETTh2 & 336/168 & GELU &  & 0.3239 & 0.3692 & 0.3264 & 0.3715 & 0.3273 & 0.3710\\
ETTh2 & 336/168 & softsign &  & 0.3303 & 0.3690 & 0.3336 & 0.3715 & 0.3308 & 0.3702\\
ETTh2 & 336/168 & tanh &  & 0.3293 & 0.3681 & 0.3334 & 0.3709 & 0.3304 & 0.3701\\
ETTh2 & 336/168 & tanh-sin &  & 0.3293 & 0.3681 & 0.3334 & 0.3709 & 0.3304 & 0.3701\\
ETTh2 & 336/336 & GELU &  & 0.3287 & 0.3847 & 0.3288 & 0.3851 & 0.3278 & 0.3833\\
ETTh2 & 336/336 & GELU+out &  & 0.3379 & 0.3867 & 0.3395 & 0.3881 & 0.3346 & 0.3831\\
ETTh2 & 336/336 & softsign &  & 0.3334 & 0.3824 & 0.3351 & 0.3839 & 0.3314 & 0.3815\\
ETTh2 & 336/336 & tanh &  & 0.3328 & 0.3817 & 0.3349 & 0.3830 & 0.3319 & 0.3814\\
ETTh2 & 336/336 & tanh+out &  & 0.3395 & 0.3856 & 0.3443 & 0.3886 & 0.3384 & 0.3857\\
ETTh2 & 336/336 & tanh-sin &  & 0.3328 & 0.3817 & 0.3349 & 0.3830 & 0.3319 & 0.3814\\
ETTh2 & 336/336 & tanh-sin+out &  & 0.3395 & 0.3856 & 0.3443 & 0.3886 & 0.3385 & 0.3857\\
ETTh2 & 336/720 & GELU &  & 0.3802 & 0.4227 & 0.3802 & 0.4229 & 0.3823 & 0.4235\\
ETTh2 & 336/720 & GELU+out &  & 0.3832 & 0.4234 & 0.3845 & 0.4241 & 0.3860 & 0.4247\\
ETTh2 & 336/720 & softsign &  & 0.3768 & 0.4173 & 0.3788 & 0.4186 & 0.3794 & 0.4188\\
ETTh2 & 336/720 & tanh &  & 0.3766 & 0.4168 & 0.3788 & 0.4180 & 0.3797 & 0.4188\\
ETTh2 & 336/720 & tanh+out &  & 0.3801 & 0.4190 & 0.3822 & 0.4201 & 0.3824 & 0.4202\\
ETTh2 & 336/720 & tanh-sin &  & 0.3766 & 0.4168 & 0.3788 & 0.4180 & 0.3797 & 0.4188\\
ETTh2 & 336/720 & tanh-sin+out &  & 0.3801 & 0.4190 & 0.3822 & 0.4201 & 0.3824 & 0.4202\\
ETTm2 & 336/96 & GELU &  & 0.1666 & 0.2555 & 0.1654 & 0.2532 & 0.1648 & 0.2545\\
ETTm2 & 336/96 & softsign &  & 0.1680 & 0.2556 & 0.1669 & 0.2533 & 0.1665 & 0.2548\\
ETTm2 & 336/96 & tanh &  & 0.1679 & 0.2553 & 0.1665 & 0.2528 & 0.1662 & 0.2540\\
ETTm2 & 336/96 & tanh-sin &  & 0.1679 & 0.2553 & 0.1665 & 0.2528 & 0.1662 & 0.2540\\
ETTm2 & 336/288 & GELU &  & 0.2615 & 0.3200 & 0.2616 & 0.3210 & 0.2651 & 0.3216\\
ETTm2 & 336/288 & GELU+out &  & 0.2672 & 0.3237 & 0.2680 & 0.3251 & 0.2700 & 0.3244\\
ETTm2 & 336/288 & softsign &  & 0.2612 & 0.3191 & 0.2619 & 0.3206 & 0.2671 & 0.3217\\
ETTm2 & 336/288 & tanh &  & 0.2612 & 0.3189 & 0.2619 & 0.3203 & 0.2672 & 0.3215\\
ETTm2 & 336/288 & tanh+out &  & 0.2667 & 0.3229 & 0.2679 & 0.3249 & 0.2702 & 0.3237\\
ETTm2 & 336/288 & tanh-sin &  & 0.2613 & 0.3189 & 0.2619 & 0.3203 & 0.2673 & 0.3215\\
ETTm2 & 336/288 & tanh-sin+out &  & 0.2667 & 0.3229 & 0.2679 & 0.3248 & 0.2702 & 0.3237\\
ETTm2 & 336/672 & GELU &  & 0.3557 & 0.3776 & 0.3556 & 0.3776 & 0.3544 & 0.3770\\
ETTm2 & 336/672 & GELU+out &  & 0.3597 & 0.3790 & 0.3582 & 0.3796 & 0.3578 & 0.3787\\
ETTm2 & 336/672 & softsign &  & 0.3594 & 0.3786 & 0.3595 & 0.3787 & 0.3582 & 0.3776\\
ETTm2 & 336/672 & tanh &  & 0.3593 & 0.3782 & 0.3593 & 0.3783 & 0.3583 & 0.3774\\
ETTm2 & 336/672 & tanh+out &  & 0.3624 & 0.3793 & 0.3616 & 0.3804 & 0.3613 & 0.3791\\
ETTm2 & 336/672 & tanh-sin &  & 0.3593 & 0.3782 & 0.3593 & 0.3783 & 0.3583 & 0.3774\\
ETTm2 & 336/672 & tanh-sin+out &  & 0.3624 & 0.3792 & 0.3616 & 0.3804 & 0.3614 & 0.3791\\
Solar2 & 96/24 & GELU &  & 0.0649 & 0.1316 & 0.0653 & 0.1316 & 0.0662 & 0.1338\\
Solar2 & 96/24 & softsign &  & 0.0651 & 0.1306 & 0.0648 & 0.1315 & 0.0656 & 0.1311\\
Solar2 & 96/24 & tanh &  & 0.0654 & 0.1309 & 0.0646 & 0.1299 & 0.0656 & 0.1318\\
Solar2 & 96/24 & tanh-sin &  & 0.0654 & 0.1310 & 0.0646 & 0.1297 & 0.0656 & 0.1316\\
Solar2 & 96/96 & GELU &  & 0.0993 & 0.1795 & 0.1000 & 0.1795 & 0.0997 & 0.1795\\
Solar2 & 96/96 & softsign &  & 0.0982 & 0.1779 & 0.0986 & 0.1771 & 0.0988 & 0.1779\\
Solar2 & 96/96 & tanh &  & 0.0983 & 0.1775 & 0.0989 & 0.1787 & 0.0995 & 0.1792\\
Solar2 & 96/96 & tanh-sin &  & 0.0983 & 0.1775 & 0.0989 & 0.1787 & 0.0995 & 0.1792\\
Solar3 & 96/24 & GELU &  & 0.2764 & 0.3014 & 0.2801 & 0.3035 & 0.2807 & 0.3043\\
Solar3 & 96/24 & softsign &  & 0.2798 & 0.3032 & 0.2833 & 0.3057 & 0.2878 & 0.3096\\
Solar3 & 96/24 & tanh &  & 0.2795 & 0.3032 & 0.2822 & 0.3036 & 0.2842 & 0.3054\\
Solar3 & 96/24 & tanh-sin &  & 0.2795 & 0.3032 & 0.2820 & 0.3036 & 0.2841 & 0.3053\\
Solar3 & 96/96 & GELU &  & 0.4478 & 0.4193 & 0.4463 & 0.4203 & 0.4452 & 0.4155\\
Solar3 & 96/96 & GELU+out &  & 0.4909 & 0.4309 & 0.4914 & 0.4294 & 0.4864 & 0.4260\\
Solar3 & 96/96 & softsign &  & 0.4506 & 0.4182 & 0.4555 & 0.4213 & 0.4528 & 0.4187\\
Solar3 & 96/96 & tanh &  & 0.4503 & 0.4181 & 0.4534 & 0.4203 & 0.4507 & 0.4179\\
Solar3 & 96/96 & tanh+out &  & 0.4956 & 0.4315 & 0.4960 & 0.4324 & 0.4898 & 0.4281\\
Solar3 & 96/96 & tanh-sin &  & 0.4503 & 0.4181 & 0.4533 & 0.4203 & 0.4506 & 0.4179\\
Solar3 & 96/96 & tanh-sin+out &  & 0.4980 & 0.4339 & 0.4959 & 0.4323 & 0.4898 & 0.4281\\
Solar5 & 96/24 & GELU &  & 0.1139 & 0.1611 & 0.1156 & 0.1636 & 0.1149 & 0.1643\\
Solar5 & 96/24 & softsign &  & 0.1134 & 0.1611 & 0.1150 & 0.1654 & 0.1129 & 0.1637\\
Solar5 & 96/24 & tanh &  & 0.1130 & 0.1604 & 0.1134 & 0.1618 & 0.1133 & 0.1642\\
Solar5 & 96/24 & tanh-sin &  & 0.1131 & 0.1604 & 0.1134 & 0.1618 & 0.1133 & 0.1641\\
Solar5 & 96/96 & GELU &  & 0.1763 & 0.2229 & 0.1762 & 0.2215 & 0.1767 & 0.2220\\
Solar5 & 96/96 & GELU+out &  & 0.2050 & 0.2363 & 0.2046 & 0.2361 & 0.2053 & 0.2368\\
Solar5 & 96/96 & softsign &  & 0.1733 & 0.2222 & 0.1737 & 0.2223 & 0.1768 & 0.2277\\
Solar5 & 96/96 & tanh &  & 0.1765 & 0.2285 & 0.1727 & 0.2187 & 0.1759 & 0.2275\\
Solar5 & 96/96 & tanh+out &  & 0.2040 & 0.2358 & 0.2039 & 0.2369 & 0.2040 & 0.2356\\
Solar5 & 96/96 & tanh-sin &  & 0.1765 & 0.2285 & 0.1727 & 0.2185 & 0.1759 & 0.2275\\
Solar5 & 96/96 & tanh-sin+out &  & 0.2040 & 0.2358 & 0.2042 & 0.2367 & 0.2040 & 0.2358\\
\multicolumn{10}{l}{\textbf{iTransformer: static regime}}\\*
ETTh2 & 96/96 & GELU &  & 0.3016 & 0.3515 & -- & -- & -- & --\\
ETTh2 & 96/96 & softsign &  & 0.3027 & 0.3505 & -- & -- & -- & --\\
ETTh2 & 96/96 & tanh &  & 0.3028 & 0.3503 & -- & -- & -- & --\\
ETTh2 & 96/96 & tanh-sin &  & 0.3028 & 0.3503 & -- & -- & -- & --\\
ETTh2 & 96/336 & GELU &  & 0.4229 & 0.4321 & -- & -- & -- & --\\
ETTh2 & 96/336 & softsign &  & 0.4283 & 0.4340 & -- & -- & -- & --\\
ETTh2 & 96/336 & tanh &  & 0.4294 & 0.4341 & -- & -- & -- & --\\
ETTh2 & 96/336 & tanh-sin &  & 0.4294 & 0.4341 & -- & -- & -- & --\\
ETTh2 & 96/720 & GELU &  & 0.4290 & 0.4472 & -- & -- & -- & --\\
ETTh2 & 96/720 & softsign &  & 0.4336 & 0.4495 & -- & -- & -- & --\\
ETTh2 & 96/720 & tanh &  & 0.4365 & 0.4506 & -- & -- & -- & --\\
ETTh2 & 96/720 & tanh-sin &  & 0.4366 & 0.4506 & -- & -- & -- & --\\
ETTm2 & 96/96 & GELU &  & 0.1842 & 0.2691 & -- & -- & -- & --\\
ETTm2 & 96/96 & softsign &  & 0.1842 & 0.2686 & -- & -- & -- & --\\
ETTm2 & 96/96 & tanh &  & 0.1839 & 0.2682 & -- & -- & -- & --\\
ETTm2 & 96/96 & tanh-sin &  & 0.1839 & 0.2682 & -- & -- & -- & --\\
ETTm2 & 96/336 & GELU &  & 0.3134 & 0.3513 & -- & -- & -- & --\\
ETTm2 & 96/336 & softsign &  & 0.3156 & 0.3526 & -- & -- & -- & --\\
ETTm2 & 96/336 & tanh &  & 0.3156 & 0.3526 & -- & -- & -- & --\\
ETTm2 & 96/336 & tanh-sin &  & 0.3156 & 0.3526 & -- & -- & -- & --\\
ETTm2 & 96/720 & GELU &  & 0.4141 & 0.4065 & -- & -- & -- & --\\
ETTm2 & 96/720 & softsign &  & 0.4155 & 0.4068 & -- & -- & -- & --\\
ETTm2 & 96/720 & tanh &  & 0.4145 & 0.4059 & -- & -- & -- & --\\
ETTm2 & 96/720 & tanh-sin &  & 0.4144 & 0.4059 & -- & -- & -- & --\\
\multicolumn{10}{l}{\textbf{iTransformer: output placement}}\\*
ETTh2 & 96/96 & GELU+out &  & 0.2997 & 0.3488 & -- & -- & -- & --\\
ETTh2 & 96/96 & ReLU &  & 0.3015 & 0.3516 & -- & -- & -- & --\\
ETTh2 & 96/96 & ReLU+out &  & 0.2997 & 0.3490 & -- & -- & -- & --\\
ETTh2 & 96/96 & softsign+out &  & 0.3016 & 0.3489 & -- & -- & -- & --\\
ETTh2 & 96/96 & Swish &  & 0.3017 & 0.3509 & -- & -- & -- & --\\
ETTh2 & 96/96 & Swish+out &  & 0.3001 & 0.3486 & -- & -- & -- & --\\
ETTh2 & 96/96 & tanh+out &  & 0.3020 & 0.3489 & -- & -- & -- & --\\
ETTh2 & 96/96 & tanh-sin+out &  & 0.3020 & 0.3489 & -- & -- & -- & --\\
ETTm2 & 96/96 & GELU+out &  & 0.1851 & 0.2695 & -- & -- & -- & --\\
ETTm2 & 96/96 & ReLU &  & 0.1843 & 0.2692 & -- & -- & -- & --\\
ETTm2 & 96/96 & ReLU+out &  & 0.1851 & 0.2696 & -- & -- & -- & --\\
ETTm2 & 96/96 & softsign+out &  & 0.1849 & 0.2687 & -- & -- & -- & --\\
ETTm2 & 96/96 & Swish &  & 0.1840 & 0.2689 & -- & -- & -- & --\\
ETTm2 & 96/96 & Swish+out &  & 0.1848 & 0.2691 & -- & -- & -- & --\\
ETTm2 & 96/96 & tanh+out &  & 0.1847 & 0.2685 & -- & -- & -- & --\\
ETTm2 & 96/96 & tanh-sin+out &  & 0.1847 & 0.2685 & -- & -- & -- & --\\
ETTm2 & 96/720 & GELU+out &  & 0.4109 & 0.4047 & -- & -- & -- & --\\
ETTm2 & 96/720 & ReLU &  & 0.4145 & 0.4066 & -- & -- & -- & --\\
ETTm2 & 96/720 & ReLU+out &  & 0.4108 & 0.4046 & -- & -- & -- & --\\
ETTm2 & 96/720 & softsign+out &  & 0.4097 & 0.4030 & -- & -- & -- & --\\
ETTm2 & 96/720 & Swish &  & 0.4141 & 0.4062 & -- & -- & -- & --\\
ETTm2 & 96/720 & Swish+out &  & 0.4104 & 0.4043 & -- & -- & -- & --\\
ETTm2 & 96/720 & tanh+out &  & 0.4085 & 0.4025 & -- & -- & -- & --\\
ETTm2 & 96/720 & tanh-sin+out &  & 0.4085 & 0.4025 & -- & -- & -- & --\\
\end{longtable}
\par\endgroup

\clearpage
\section{Offline Model, Reference Prompts, and Reproduction}
\label{app:template_induction_protocol}

This appendix provides the model configuration, recovered task instruction, reusable candidate-generation prompt, and executable materials for reproducing the activation policy. The original Chinese task instruction and its English translation accompany the source package, together with the evaluated JSON, calibration tables, verifier, and expected outputs.

\subsection{Model and Artifact Provenance}
\label{app:model_provenance}

The offline assistant was OpenAI Codex using the recorded model identifier \texttt{gpt-5.5} and reasoning effort \texttt{xhigh}. The session metadata records Codex CLI version \texttt{0.135.0-alpha.1}. On 5 June 2026, the task instruction was issued at 14:32:23 UTC, the v2 template file was created at 14:44:33 UTC, and three extrapolation templates were revised at 14:53:46 UTC. Applying the recovered creation and revision patches reproduces the evaluated JSON byte for byte, with SHA-256 prefix \texttt{67305b22}. The supplied \texttt{recovered\_provenance.json} records the event timestamps, model settings, and complete checksums.

The task was executed interactively with access to prior project evidence. The historical instruction explicitly requests Traffic-horizon repairs, and the evaluated JSON marks the three corresponding templates as extrapolation hypotheses. Thus the reported study evaluates a frozen policy under calibrated leave-dataset validation; the no-extrapolation comparison separates those repairs from the remaining rules. The session records the model identifier and reasoning effort, but does not specify a separate dated backend snapshot, temperature, top-$p$, or sampling seed. Candidate regeneration uses the documented model configuration and reference prompt; exact numerical reproduction uses the frozen JSON and verifier.

\subsection{Recovered Historical Task Instruction}
\label{app:historical_prompt}

The following is an English translation of the recovered task-specific user instruction. It preserves the requested benchmark scope, rule changes, evaluation criteria, and assumptions. The original Chinese text is provided without rewriting in \texttt{historical\_task\_prompt\_zh.md}. This instruction documents the interactive development task; the standalone reference prompt in the next subsection specifies the reusable candidate-generation interface.

\begin{lstlisting}[style=referenceprompt]
PLEASE IMPLEMENT THIS PLAN:
Phase-9 Plan: Validate and Improve the LLM-Assisted Calibrated Router

SUMMARY
The goal is to validate and improve the LLM-assisted calibrated factor-router.
The main results concern 17 Informer datasets / 51 dataset-horizon cells.
PatchTST and iTransformer are architecture-dependence diagnostics and are not
pooled into the main claim. The user has prioritized repairing the Traffic
horizon switch: Traffic-96/192 should abstain to GELU, whereas Traffic-336
should trigger a bounded static action such as softsign or tanh_sin.

This phase does not pursue pure zero-shot success. An acceptable outcome under
leave-dataset calibration is a small positive gain over gelu_all, no material
regression relative to the train-fold static baseline, and lower oracle regret.

KEY CHANGES
1. Establish a unified Phase-9 evaluation protocol.
   - Main evidence: Informer Phase-4 + Phase-5 + Phase-8, 17 datasets / 51 cells.
   - Hold out each target dataset; train/calibrate using completed Informer
     results from all other datasets.
   - Flag the one-seed Electricity/Traffic probes separately and provide a
     three-seed-only sensitivity table.
2. Extend the router evaluator.
   - Support arbitrary leave_dataset targets, rather than only the previous
     Weather/Exchange/ILI or Electricity/Traffic groups.
   - Report gain versus GELU and the train-fold strongest static baseline,
     oracle regret, capture ratio, abstention, and configuration/action hits.
   - gelu_all is the deployable default; train_static is selected exclusively
     from the training fold; oracle is the target cell's lowest-MSE action,
     including GELU, used only as an upper bound.
3. Improve the LLM-assisted frozen templates.
   - Retain existing templates such as decoder_tanh_low_volatility.
   - Add short_high_dim_gelu_guard: abstain to GELU for high-dimensional,
     traffic-like tasks at short or medium horizons.
   - Add long_horizon_softsign_candidate: trigger softsign_all for
     high-dimensional traffic-like tasks at long horizons.
   - Add long_horizon_bounded_static_candidate: at long horizons with historical
     support for bounded static actions, choose softsign/tanh_sin using the
     higher training-fold uplift.
   - Serialize every template as frozen JSON. The LLM proposes templates
     offline; only the deterministic verifier selects and executes at test time.
4. Record the Phase-9 artifacts.
   - .aris/phase9_calibrated_router_plan_20260605.md
   - .aris/phase9_router_templates_v2_20260605.json
   - .aris/phase9_leave_dataset_router_eval_20260605/
   - Update the README and paper story after completed results establish the
     final numerical values.

TEST PLAN
1. CPU-only verification.
   - Reconfirm Phase-4 972/972, Phase-5 189/189, and Phase-8 42/42 runs.
   - Run leave-dataset evaluation without launching training.
   - Inspect each dataset's predictions and the Traffic 96/192/336 decisions.
2. Router success criteria.
   - GO: positive mean gain over GELU; nonnegative or a small explainable
     negative gain versus train_static; lower regret than abstain-all-GELU;
     Traffic-336 no longer incorrectly abstains.
   - CALIBRATE: small but unstable positive gain over GELU; retain calibrated
     routing and add targeted seeds.
   - PIVOT: if gains come only from abstention and non-GELU oracle choices are
     missed, describe the method as factor-guided low-cost activation search.
3. Required result tables.
   - A: 17-dataset leave-dataset router summary.
   - B: per-dataset predictions highlighting the Traffic horizon switch.
   - C: no-LLM versus LLM-assisted versus abstain-only versus train-static.
   - Appendix: PatchTST/iTransformer transfer probes; do not claim universal
     transfer from weak cross-architecture effects.
4. If GO or CALIBRATE is reached, launch targeted remote multi-seed checks.
   - Traffic-336: gelu_all, softsign_all, tanh_sin001_all.
   - Traffic-96/192: gelu_all and the router-selected non-GELU candidate.
   - Electricity-96/336: gelu_all and enc_gelu_dec_tanh.
   - Additional seeds: 2025 and 2026. Do not train on the local machine.

ASSUMPTIONS
The main claim concerns Informer; PatchTST/iTransformer establish
architecture-dependent sensitivity. Small gains are positive signals but must
be accompanied by regret, abstention, and per-dataset outcomes. Phase-9 is
internal calibrated validation with iterative improvement, not strict
zero-shot evaluation on entirely new datasets. A stronger claim would require
freezing the v2 router and then evaluating on a new clean held-out dataset.
\end{lstlisting}

\subsection{Standalone Reference Prompt}
\label{app:reference_prompt}

Use the following complete reference prompt with the exported calibration packet and \texttt{rule\_template.schema.json}, selecting \texttt{gpt-5.5} with reasoning effort \texttt{xhigh}. The packet contains input-derived factors and calibration-side action responses, after removing the declared held-out dataset. Save the returned JSON and validate it before evaluation. The prompt below is the normalized reusable protocol, while the preceding subsection preserves the historical development instruction.

\begin{lstlisting}[style=referenceprompt]
ROLE
You are an offline candidate-rule author. You do not make a forecasting-time
decision. The deterministic verifier is the only component allowed to accept a
rule or select an activation action.

INPUT
You receive a hash-recorded calibration packet containing:
1. input-derived factors and task metadata;
2. calibration-side action/comparator names and continuous MSE uplift;
3. calibration-side guard labels such as whether the default is oracle; and
4. the activation action vocabulary.

The packet has already removed every row from the declared held-out dataset.
Do not infer, request, or fabricate held-out outcomes. Do not use target-cell
MSE, oracle action, or post-decision evaluation metrics.

TASK
Propose a small set of interpretable factor--action hypotheses. Each ordinary
rule should contain one to five conditions and use calibration-fold quantiles
instead of hard-coded numeric thresholds. A rule must identify its action name,
action family, condition-combination mode, and a short mechanism/failure
hypothesis. Prefer a compact portfolio that covers distinct regimes or
placements over many near-duplicate rules.

An extrapolation hypothesis is permitted only when explicitly marked with
"allow_extrapolation": true. It will be reported separately and must not be
described as training-supported zero-shot evidence.

OUTPUT
Return one JSON object and no prose. The object must validate against
rule_template.schema.json. Do not report that a candidate is accepted, because
support, precision, mean uplift, thresholds, and final eligibility are computed
later by the deterministic verifier on each calibration fold.
\end{lstlisting}

\subsection{Execution and Verification}
\label{app:replay}

The source package includes a self-contained \texttt{artifacts/rule\_induction/} directory. Run the following commands from that directory. The first command installs the tested dependency versions; the second refits fold-local thresholds and reproduces the recorded decisions from the bundled factors, costs, and templates.

\begin{lstlisting}[style=referenceprompt]
python -m pip install -r requirements.txt
python replay.py --output-dir replay_output
\end{lstlisting}

The replay was checked with Python 3.12.7, NumPy 2.3.2, pandas 2.3.2, and PyTorch 2.9.1. It verifies file checksums, compares all 408 decisions across four confidence settings and two fallback choices, and checks their errors and gains against the frozen outputs. The selected balanced/CF-SBC setting contains 51 tasks and reproduces the reported 7.86\% mean MSE reduction, 0.89\% increment over CF-SBC, and 4.82\% oracle regret. It requires neither GPU training nor an LLM call.

The verifier uses an uplift margin of 0.002, minimum pair count 10, default rule support 4 and precision 0.55, and guard support 2 and precision 0.60. Individual templates may override the defaults, as recorded in the supplied JSON. The balanced action filter requires objective at least 0.002, precision at least 0.65, and mean uplift at least zero. Quantiles, CF-SBC, and within-family rankings are fitted after excluding the target dataset.

To generate a fresh calibration packet for an example held-out dataset and validate a new candidate set, use:

\begin{lstlisting}[style=referenceprompt]
python code/export_rule_induction_packet.py --heldout ETTh1 
  --cells data/phase9_unified_cells.csv 
  --pairs data/phase9_unified_pairs.csv 
  --output-dir packet_etth1
python code/validate_rule_templates.py candidate_rules.json
\end{lstlisting}

The first command is entered as one line (or with shell line continuations). Attach the two packet CSVs, packet manifest, and JSON schema to the reference prompt. Candidate outputs are frozen and evaluated as a new policy version. The historical results remain tied to \texttt{evaluated\_templates.json}; regenerating candidates does not silently replace that evaluated artifact.

\clearpage
\section{Dataset-Level Policy Results}
\label{app:dataset_results}

Table~\ref{tab:phase9_per_dataset_summary} resolves the main policy result into its 17 dataset variants. It uses the same balanced confidence setting, CF-SBC fallback, and 51 tasks as the main comparison. Gain, regret, and alignment are averaged within each dataset; the overall result weights dataset--horizon tasks rather than dataset rows equally. The table makes horizon-dependent behavior and differences between dataset families visible alongside the aggregate improvement.

\begin{table}[!htbp]
\centering\footnotesize
\caption{Dataset-level results for the balanced policy with CF-SBC fallback.}
\label{tab:phase9_per_dataset_summary}
\setlength{\tabcolsep}{3pt}
\renewcommand{\arraystretch}{1.05}
\begin{tabular*}{\linewidth}{@{\extracolsep{\fill}}lrrrrrrr@{}}
\toprule
Dataset & Tasks & \shortstack{Gain vs\\GELU} & \shortstack{Gain vs\\CF-SBC} & Regret & Positive & \shortstack{Config.\\align.} & \shortstack{Family\\align.}\\\midrule
ETTh1 & 5 & +3.35\% & -3.57\% & 6.89\% & 60.0\% & 0.0\% & 80.0\%\\
ETTh2 & 5 & +32.73\% & +0.00\% & 3.74\% & 100.0\% & 40.0\% & 60.0\%\\
ETTm1 & 5 & +6.43\% & +0.00\% & 4.23\% & 80.0\% & 20.0\% & 100.0\%\\
ETTm2 & 5 & +37.12\% & +0.00\% & 6.60\% & 100.0\% & 20.0\% & 60.0\%\\
Electricity & 3 & +0.24\% & +3.62\% & 0.58\% & 33.3\% & 66.7\% & 100.0\%\\
Exchange & 3 & -15.50\% & +0.00\% & 17.68\% & 0.0\% & 0.0\% & 100.0\%\\
ILI & 3 & +2.11\% & -0.92\% & 1.09\% & 100.0\% & 33.3\% & 100.0\%\\
Solar1 & 2 & -0.21\% & +0.00\% & 2.47\% & 50.0\% & 0.0\% & 50.0\%\\
Solar2 & 2 & +7.76\% & +0.00\% & 0.99\% & 100.0\% & 0.0\% & 100.0\%\\
Solar3 & 2 & -1.85\% & +0.00\% & 1.85\% & 0.0\% & 0.0\% & 0.0\%\\
Solar4 & 2 & +1.17\% & +0.00\% & 0.79\% & 100.0\% & 50.0\% & 100.0\%\\
Solar5 & 2 & +4.66\% & +0.00\% & 1.31\% & 100.0\% & 0.0\% & 50.0\%\\
Solar6 & 2 & +1.70\% & +0.00\% & 0.00\% & 100.0\% & 100.0\% & 100.0\%\\
Solar7 & 2 & +1.46\% & +0.00\% & 32.31\% & 50.0\% & 0.0\% & 50.0\%\\
Solar8 & 2 & +1.14\% & +0.00\% & 0.55\% & 100.0\% & 50.0\% & 100.0\%\\
Traffic & 3 & +1.33\% & +1.36\% & 0.00\% & 33.3\% & 100.0\% & 100.0\%\\
Weather & 3 & +2.16\% & +17.00\% & 0.00\% & 100.0\% & 100.0\% & 100.0\%\\
\bottomrule
\end{tabular*}
\par\smallskip\begin{minipage}{\linewidth}\footnotesize Positive denotes the fraction of tasks improved over GELU; configuration and family alignment compare the selected action with the evaluation oracle.\end{minipage}
\end{table}

\FloatBarrier

\section{Architecture-Specific Activation Comparisons}
\label{app:architecture_results}

The transfer experiments test how activation choices interact with backbone topology. Tables~\ref{tab:patchtst_transfer} and~\ref{tab:itransformer_transfer} compare hidden-FFN activation regimes. Tables~\ref{tab:patchtst_output_tanh} and~\ref{tab:itransformer_placeprobe} examine output placement. These summaries use the same runs listed in Table~\ref{tab:all_seed_results}; they retain each architecture's training protocol and seed coverage. The main selector evaluation concerns Informer, while these controlled comparisons identify the configuration choices that need to be calibrated for each backbone.

\begin{table}[!htbp]
\centering\footnotesize
\caption{PatchTST hidden-FFN activation comparisons.}
\label{tab:patchtst_transfer}
\setlength{\tabcolsep}{3pt}
\renewcommand{\arraystretch}{1.05}
\begin{tabular*}{\linewidth}{@{\extracolsep{\fill}}lrrrlr@{}}
\toprule
Dataset & Horizon & Seeds & GELU MSE & Best regime & Gain\\\midrule
ETTh2 & 168 & 3 & 0.325880 & GELU & 0.00\%\\
ETTh2 & 336 & 3 & 0.328450 & GELU & 0.00\%\\
ETTh2 & 720 & 3 & 0.380881 & softsign & 0.67\%\\
ETTm2 & 96 & 3 & 0.165589 & GELU & 0.00\%\\
ETTm2 & 288 & 3 & 0.262739 & GELU & 0.00\%\\
ETTm2 & 672 & 3 & 0.355232 & GELU & 0.00\%\\
Solar2 & 24 & 3 & 0.065450 & tanh-sin & 0.42\%\\
Solar2 & 96 & 3 & 0.099666 & softsign & 1.12\%\\
Solar3 & 24 & 3 & 0.279035 & GELU & 0.00\%\\
Solar3 & 96 & 3 & 0.446409 & GELU & 0.00\%\\
Solar5 & 24 & 3 & 0.114792 & tanh-sin & 1.33\%\\
Solar5 & 96 & 3 & 0.176366 & softsign & 1.01\%\\
\bottomrule
\end{tabular*}
\par\smallskip\begin{minipage}{\linewidth}\footnotesize MSE is averaged over the listed seeds. Gain compares the best evaluated regime with the matched GELU baseline; all configurations retain a linear output head.\end{minipage}
\end{table}

\begin{table}[!htbp]
\centering\footnotesize
\caption{PatchTST output-placement comparisons.}
\label{tab:patchtst_output_tanh}
\setlength{\tabcolsep}{3pt}
\renewcommand{\arraystretch}{1.05}
\begin{tabular*}{\linewidth}{@{\extracolsep{\fill}}lrrr@{}}
\toprule
Hidden activation & Paired runs & Win rate & Mean gain\\\midrule
tanh & 18 & 0.0\% & -5.28\%\\
tanh-sin & 18 & 0.0\% & -5.31\%\\
GELU & 18 & 0.0\% & -5.44\%\\
\bottomrule
\end{tabular*}
\par\smallskip\begin{minipage}{\linewidth}\footnotesize Each pair uses the same task, seed, and hidden activation. Gain is the relative MSE reduction from adding output tanh to the linear-head configuration; positive values favor output tanh.\end{minipage}
\end{table}

\begin{table}[!htbp]
\centering\footnotesize
\caption{iTransformer hidden-FFN activation comparisons.}
\label{tab:itransformer_transfer}
\setlength{\tabcolsep}{3pt}
\renewcommand{\arraystretch}{1.05}
\begin{tabular*}{\linewidth}{@{\extracolsep{\fill}}lrrrlr@{}}
\toprule
Dataset & Horizon & Seeds & GELU MSE & Best regime & Gain\\\midrule
ETTh2 & 96 & 1 & 0.301598 & GELU & 0.00\%\\
ETTh2 & 336 & 1 & 0.422894 & GELU & 0.00\%\\
ETTh2 & 720 & 1 & 0.428986 & GELU & 0.00\%\\
ETTm2 & 96 & 1 & 0.184226 & tanh-sin & 0.19\%\\
ETTm2 & 336 & 1 & 0.313418 & GELU & 0.00\%\\
ETTm2 & 720 & 1 & 0.414092 & GELU & 0.00\%\\
\bottomrule
\end{tabular*}
\par\smallskip\begin{minipage}{\linewidth}\footnotesize MSE is averaged over the listed seeds. Gain compares the best evaluated regime with the matched GELU baseline; all configurations retain a linear output head.\end{minipage}
\end{table}

\begin{table}[!htbp]
\centering\footnotesize
\caption{iTransformer mixed-configuration probe: hidden activations and output placement.}
\label{tab:itransformer_placeprobe}
\setlength{\tabcolsep}{3pt}
\renewcommand{\arraystretch}{1.05}
\begin{tabular*}{\linewidth}{@{\extracolsep{\fill}}lrlr@{}}
\toprule
Dataset & Horizon & Best configuration & Gain vs GELU\\\midrule
ETTh2 & 96 & GELU+out & +0.64\%\\
ETTm2 & 96 & tanh-sin & +0.19\%\\
ETTm2 & 720 & tanh+out & +1.35\%\\
\bottomrule
\end{tabular*}
\par\smallskip\begin{minipage}{\linewidth}\footnotesize Each case uses seed 2024. The candidate portfolio includes hidden-FFN alternatives and output-tanh interventions. The suffix +out denotes output tanh; configurations without that suffix retain a linear output head. Two winning configurations use output tanh; the ETTm2/96 winner changes only the hidden activation.\end{minipage}
\end{table}

\FloatBarrier

\section{Dataset Sources and Training Protocols}
\label{app:data_protocol}

This appendix identifies the data used in the experiments and the settings needed to repeat each within-task comparison. Table~\ref{tab:dataset_snapshots} reports the actual input-file dimensions and split sizes. The accompanying \texttt{artifacts/data\_protocol/dataset\_manifest.json} records filenames, timestamp ranges, column names, missing-value counts, and SHA-256 hashes for all 17 snapshots. These hashes specify the experimental files; the historical download dates and upstream release tags were not retained.

\subsection{Original sources and benchmark versions}

\textbf{ETT.} ETTh1, ETTh2, ETTm1, and ETTm2 are the electricity-transformer datasets introduced with Informer \citep{zhou2021informer}. We use the seven-channel CSV files distributed through \url{https://github.com/zhouhaoyi/ETDataset}. ETTh samples are hourly and ETTm samples are 15-minute observations. The loaders use the first 12, next 4, and following 4 30-day-equivalent months for training, validation, and testing, respectively; later ETT rows are unused.

\textbf{Weather.} The observations originate from the weather station of the Max Planck Institute for Biogeochemistry in Jena \citep{weatherjena}. We use the 21-channel, 10-minute \texttt{weather.csv} benchmark associated with Autoformer \citep{wu2021autoformer}, rather than downloading a newly updated station archive. The station archive is available at \url{https://www.bgc-jena.mpg.de/wetter/}; the prepared benchmark collection is linked from \url{https://github.com/thuml/Autoformer}.

\textbf{Electricity, Traffic, and Exchange.} The Electricity source is the UCI ElectricityLoadDiagrams20112014 collection \citep{trindade2015electricity}; our benchmark contains 321 hourly consumption channels. Traffic contains 862 hourly road-occupancy channels derived from the California Department of Transportation PeMS archive \citep{caltranspems}. Exchange contains eight daily exchange-rate channels. Their forecasting-benchmark provenance is documented by LSTNet \citep{lai2018lstnet} and its data repository, \url{https://github.com/laiguokun/multivariate-time-series-data}. We use the timestamped CSV forms adopted in the Autoformer benchmark collection \citep{wu2021autoformer}; the local-file hashes distinguish these prepared inputs from the larger original archives.

\textbf{ILI.} The seven-channel weekly national influenza-like-illness benchmark is derived from CDC ILINet surveillance \citep{cdcfluview} and distributed as \texttt{national\_illness.csv} in the Autoformer benchmark collection \citep{wu2021autoformer}. These are aggregate surveillance series, not individual patient records. The original surveillance documentation and access portal are available at \url{https://www.cdc.gov/fluview/overview/}.

\textbf{Solar.} Solar1--Solar8 are the eight photovoltaic stations in the Chinese State Grid Renewable Energy Generation Forecasting Competition dataset described by Chen and Xu \citep{chen2022solar}. The public archive is Figshare version 4 \citep{chen2022solardata}, with processed workbooks mirrored by the data authors at \url{https://github.com/Bob05757/Renewable-energy-generation-input-feature-variables-analysis}. The source collection covers 2019--2020 at 15-minute resolution; Solar3 contains January--July 2019, and Solar8 retains the source's shorter record with timestamp gaps. Each experimental CSV contains three irradiance channels, atmospheric pressure, power, and a sixth meteorological channel: air temperature for seven sites and relative humidity for Solar3. Solar3's historical CSV calls that humidity column \texttt{Temperature}; the source mapping identifies its physical meaning. Conversion renames and reorders columns, clips negative power to zero, and fills six isolated Solar4 power gaps by neighboring-value linear interpolation (two nighttime zeros and four daytime values). The supplied \texttt{reconstruct\_solar.py} reproduces every input value and timestamp from the processed workbooks; the retained Solar4 repair record documents the six replacements. Station capacities, in order, are 50, 130, 30, 130, 110, 35, 30, and 30 MW. The source-check record in the accompanying package identifies the mirrored workbooks and their fixed repository revision.

\begin{table}[!htbp]
\centering\small
\caption{Experimental dataset snapshots. Channels exclude the timestamp. Split sizes count target-region rows; validation and test windows also use the preceding input context.}
\label{tab:dataset_snapshots}
\setlength{\tabcolsep}{3pt}
\begin{tabular*}{\linewidth}{@{\extracolsep{\fill}}lrrrrr@{}}
\toprule Dataset & Rows & Channels & Train & Validation & Test\\\midrule
ETTh1 & 17,420 & 7 & 8,640 & 2,880 & 2,880\\
ETTh2 & 17,420 & 7 & 8,640 & 2,880 & 2,880\\
ETTm1 & 69,680 & 7 & 34,560 & 11,520 & 11,520\\
ETTm2 & 69,680 & 7 & 34,560 & 11,520 & 11,520\\
Solar1 & 70,176 & 6 & 49,123 & 7,018 & 14,035\\
Solar2 & 70,176 & 6 & 49,123 & 7,018 & 14,035\\
Solar3 & 20,352 & 6 & 14,246 & 2,036 & 4,070\\
Solar4 & 70,176 & 6 & 49,123 & 7,018 & 14,035\\
Solar5 & 70,176 & 6 & 49,123 & 7,018 & 14,035\\
Solar6 & 70,176 & 6 & 49,123 & 7,018 & 14,035\\
Solar7 & 70,176 & 6 & 49,123 & 7,018 & 14,035\\
Solar8 & 69,408 & 6 & 48,585 & 6,942 & 13,881\\
Weather & 52,696 & 21 & 36,887 & 5,270 & 10,539\\
Exchange & 7,588 & 8 & 5,311 & 760 & 1,517\\
ILI & 966 & 7 & 676 & 97 & 193\\
Electricity & 26,304 & 321 & 18,412 & 2,632 & 5,260\\
Traffic & 17,544 & 862 & 12,280 & 1,756 & 3,508\\
\bottomrule\end{tabular*}
\end{table}

\subsection{Preprocessing and evaluation}

All tasks use multivariate-to-multivariate prediction. The six non-time Solar variables, for example, are both input and forecast channels; reported Solar MSE and MAE average over all six channels rather than power alone. The supplied CSVs contain no missing entries, and the loaders apply no additional imputation. Each channel is standardized using its training-partition mean and standard deviation. The validation and test target regions are chronologically disjoint from training; their input windows may include the preceding $L$ observations as context. ETT uses the fixed boundaries above. Other datasets use $\lfloor0.7N\rfloor$ training rows and $\lfloor0.2N\rfloor$ test rows, with the remaining rows assigned to validation.

MSE and MAE are computed on the standardized scale, averaging over evaluated windows, forecast steps, and channels. Test windows are not shuffled. The evaluated loaders retain complete batches and omit the final incomplete batch. Model selection uses validation MSE with early stopping and restores the best checkpoint for test evaluation. Seeds 2024, 2025, and 2026 are reported wherever available in Appendix~\ref{app:complete_results}; the iTransformer probes and the initial Electricity/Traffic stress block use seed 2024. Comparisons pair the same dataset, horizon, seed, and architecture protocol.

\subsection{Training settings}
\label{app:training_protocol}

\textbf{Informer.} We use Adam with initial learning rate $10^{-4}$, MSE loss, at most six epochs, patience three, and batch size 32. The learning-rate schedule is the implementation's \texttt{type1} schedule, which sets the rate after epoch $e$ to $10^{-4}2^{-(e-1)}$. The model uses ProbSparse attention, time-feature embeddings, $d_{\mathrm{model}}=512$, $d_{\mathrm{ff}}=2048$, eight heads, dropout 0.05, encoder distillation, and zero padding for the unknown decoder horizon. Table~\ref{tab:informer_protocol} supplies the horizon-specific input lengths, decoder context lengths, stack depths, and attention factors. In particular, ILI uses $L/K=36/18$, and Weather, Exchange, Electricity, and Traffic use $96/48$.

\begingroup\small
\begin{longtable}{@{}lrrrrrr@{}}
\caption{Informer task-specific settings. $H$ is horizon, $L$ is input length, and $K$ is decoder context length. Other settings are given in Appendix~\ref{app:training_protocol}.}\label{tab:informer_protocol}\\
\toprule Dataset & $H$ & $L$ & $K$ & Encoder layers & Decoder layers & Factor\\\midrule\endfirsthead
\caption[]{Informer task-specific settings (continued)}\\\toprule Dataset & $H$ & $L$ & $K$ & Encoder layers & Decoder layers & Factor\\\midrule\endhead
\bottomrule\endlastfoot
ETTh1 & 24 & 48 & 48 & 2 & 1 & 3\\
ETTh1 & 48 & 96 & 48 & 2 & 1 & 5\\
ETTh1 & 168 & 168 & 168 & 2 & 1 & 5\\
ETTh1 & 336 & 168 & 168 & 2 & 1 & 5\\
ETTh1 & 720 & 336 & 336 & 2 & 1 & 5\\
ETTh2 & 24 & 48 & 48 & 2 & 1 & 5\\
ETTh2 & 48 & 96 & 96 & 2 & 1 & 5\\
ETTh2 & 168 & 336 & 336 & 3 & 2 & 5\\
ETTh2 & 336 & 336 & 168 & 3 & 2 & 5\\
ETTh2 & 720 & 720 & 336 & 3 & 2 & 5\\
ETTm1 & 24 & 672 & 96 & 2 & 1 & 5\\
ETTm1 & 48 & 96 & 48 & 2 & 1 & 5\\
ETTm1 & 96 & 384 & 384 & 2 & 1 & 5\\
ETTm1 & 288 & 672 & 288 & 2 & 1 & 5\\
ETTm1 & 672 & 672 & 384 & 2 & 1 & 5\\
ETTm2 & 24 & 672 & 96 & 2 & 1 & 5\\
ETTm2 & 48 & 96 & 48 & 2 & 1 & 5\\
ETTm2 & 96 & 384 & 384 & 2 & 1 & 5\\
ETTm2 & 288 & 672 & 288 & 2 & 1 & 5\\
ETTm2 & 672 & 672 & 384 & 2 & 1 & 5\\
Solar1--Solar8 & 24 & 96 & 48 & 2 & 1 & 5\\
Solar1--Solar8 & 96 & 96 & 48 & 2 & 1 & 5\\
Weather & 96 & 96 & 48 & 2 & 1 & 5\\
Weather & 192 & 96 & 48 & 2 & 1 & 5\\
Weather & 336 & 96 & 48 & 2 & 1 & 5\\
Exchange & 96 & 96 & 48 & 2 & 1 & 5\\
Exchange & 192 & 96 & 48 & 2 & 1 & 5\\
Exchange & 336 & 96 & 48 & 2 & 1 & 5\\
ILI & 24 & 36 & 18 & 2 & 1 & 5\\
ILI & 36 & 36 & 18 & 2 & 1 & 5\\
ILI & 48 & 36 & 18 & 2 & 1 & 5\\
Electricity & 96 & 96 & 48 & 2 & 1 & 5\\
Electricity & 192 & 96 & 48 & 2 & 1 & 5\\
Electricity & 336 & 96 & 48 & 2 & 1 & 5\\
Traffic & 96 & 96 & 48 & 2 & 1 & 5\\
Traffic & 192 & 96 & 48 & 2 & 1 & 5\\
Traffic & 336 & 96 & 48 & 2 & 1 & 5\\
\end{longtable}

\endgroup

\textbf{PatchTST.} The probes use Adam, MSE loss, learning rate $10^{-4}$, at most 20 epochs, patience five, batch size 128, three encoder layers, patch length 16, stride eight, and reversible instance normalization. ETTh2 uses $L=336$, model width 16, four heads, FFN width 128, and dropout 0.3. ETTm2 uses $L=336$, width 128, 16 heads, FFN width 256, and dropout 0.2. Solar2, Solar3, and Solar5 use $L=96$, width 128, eight heads, FFN width 256, and dropout 0.2. Head dropout equals the listed dropout. ETTh2 uses the implementation's \texttt{type3} schedule; ETTm2 and Solar use the \texttt{TST} one-cycle schedule with warm-up fraction 0.4. The accompanying protocol-source files preserve the exact schedule definitions.

\textbf{iTransformer.} The probes use Adam, MSE loss, learning rate $10^{-4}$, at most ten epochs, patience three, batch size 32, $L=96$, two encoder layers, width 128, eight heads, FFN width 128, dropout 0.1, and the \texttt{type1} learning-rate schedule. The implementation retains its input normalization and projection head. Each activation comparison uses identical settings apart from its hidden or output nonlinearity.

The file \texttt{training\_runs.csv} in \texttt{artifacts/data\_protocol/} joins all 1,471 reported runs to the retained protocol records and checks both MSE and MAE against the full-results CSV. A separate \texttt{training\_protocols.csv} lists unique settings. The frozen loaders and configuration scripts document exact split, normalization, scheduling, and batching behavior; the source-package manifest provides file-level version identifiers.
\FloatBarrier

\reftitle{References}
\bibliography{references}
\end{document}